\documentclass[lettersize,journal]{IEEEtran}
\usepackage{amsmath,amsfonts}
\usepackage{array}
\usepackage[caption=false,font=normalsize,labelfont=sf,textfont=sf]{subfig}
\usepackage{textcomp}
\usepackage{stfloats}
\usepackage{url}
\usepackage{verbatim}
\usepackage{graphicx}
\def\BibTeX{{\rm B\kern-.05em{\sc i\kern-.025em b}\kern-.08em
    T\kern-.1667em\lower.7ex\hbox{E}\kern-.125emX}}
\usepackage{balance}

\usepackage{xspace}
\newcommand{\Design}{\textit{$\mathsf{CRUISE}$\xspace}}

\usepackage{algorithm}
\usepackage{amsmath}
\usepackage{algorithm}
\usepackage{algpseudocode}
\usepackage{makecell}
\usepackage{booktabs}
\usepackage{multirow}
\usepackage{colortbl} 
\usepackage{xcolor}
\usepackage{pgfplots}
\pgfplotsset{compat=1.18}

\usepackage{pifont}
\newcommand{\cmark}{\ding{51}}%
\newcommand{\xmark}{\ding{55}}%
\definecolor{lightblue}{RGB}{173, 216, 230}
\definecolor{lavender}{RGB}{230, 230, 250}
\definecolor{palered}{RGB}{255, 210, 210}
\usepackage{enumitem}

\begin{document}

\title{CRUISE: Vision-Language Model-Guided Uncertainty-Aware Cross-Modal Sensor Fusion for Robust Autonomous Driving}

\author{Junyao Wang \textit{(Graduate Student Member, IEEE)}, Yulin Xu \textit{(Graduate Student Member, IEEE)}, Yu Li \textit{(Graduate Student Member, IEEE)}, Praneet Iddamsetty  \textit{(Graduate Student Member, IEEE)}, 
Jonathon Smereka \textit{(US Army, GVSC)}, Pramod Khargonekar \textit{(Fellow, IEEE)}, Mohammad Abdullah Al Faruque \textit{(Fellow, IEEE)}
\thanks{Junyao Wang is with the Department of Computer Science, University of California, Irvine, 92697, USA. Email: junyaow4@uci.edu.\\Yulin Xu, Yu Li, Praneet Iddamsetty, and Professor Pramod Khargonekar are with the Department of Electrical Engineering and Computer Science, University of California, Irvine, 92697, USA. Email: \{yulinx8, yul79, pramod.khargonekar\}@uci.edu.\\ Dr. Jonathon Smereka is the Ground Vehicle Robotics (GVR) Senior Technical Expert for research at the U.S. Army DEVCOM Ground Vehicle Systems Center (GVSC). \\
Professor Mohammad Abdullah Al Faruque is with the Department of Computer Science, the Department of Electrical Engineering and Computer Science, and the Department of Mechanical and Aerospace Engineering, University of California, Irvine, 92697, USA. Email: alfaruqu@uci.edu}
\vspace{-5mm}}

\markboth{Journal of \LaTeX\ Class Files,~Vol.~18, No.~9, September~2020}%
{How to Use the IEEEtran \LaTeX \ Templates}

\maketitle

\begin{abstract}
Modern autonomous vehicles are equipped with multiple sensors, such as cameras, LiDAR, and radar, for comprehensive environmental perception. However, robust cross-modal feature fusion remains a critical challenge, as the reliability of each sensor varies significantly across diverse real-world driving conditions, including poor visibility and adverse weather.
While uncertainty quantification (UQ) mitigates this issue by allowing models to prioritize reliable signals, existing uncertainty-aware fusion methods typically rely on simple feature-level uncertainty estimates and thus often fail to generalize effectively in complex, out-of-distribution scenarios. 
To address this limitation, we propose CRUISE, a novel uncertainty-aware cross-modal sensor fusion framework. CRUISE integrates a vision-language model (VLM)-guided UQ module that generates fine-grained, pixel-level uncertainty estimates. By leveraging the VLM’s rich prior knowledge and superior contextual reasoning, our approach provides a highly informative guide for the fusion process. Furthermore, we introduce a dynamic adaptation mechanism that explicitly models and captures cross-modal dependencies, ensuring the framework fully exploits the inherent complementary nature of multi-sensor inputs.  
\textcolor{black}{Experiments on aiMotive for 3D object detection and on DeLiVER and MFNet for semantic segmentation show that CRUISE achieves the highest mean performance among the compared state-of-the-art (SOTA) sensor-fusion methods. 
Under the evaluated sensor corruptions, CRUISE also outperforms the strongest competing UQ methods, with relative improvements of 2.99\% in mean all-point AP on aiMotive and 4.25\% and 1.56\% in mIoU on the DeLiVER test set and MFNet, respectively.}
\end{abstract}
    
\section{Introduction}\label{sec:intro}

Modern autonomous vehicles (AVs) employ sensor fusion to integrate data collected by diverse sensors, such as cameras, LiDAR, and radar, to achieve comprehensive perception of their surroundings~\cite{wang2024rs2g,fayyad2020deep}. 
However, sensor reliability varies significantly across different scenarios, introducing uncertainty into perception and 
posing substantial challenges for robust feature fusion. 
For instance, while cameras capture rich visual details, they are susceptible to lighting variations and occlusions. Conversely, LiDAR and radar offer greater resilience against visual perturbations but are susceptible to multi-path interference and signal disruptions~\cite{bhupathiraju2023emi,li2020lidar, chen2023futr3d}. The dynamic nature of real-world driving means these reliability shifts are constantly present, demanding systems that can adapt on the fly. 
To address the challenge of unreliable sensor inputs, uncertainty quantification (UQ) is employed to estimate the confidence of each sensor modality. This mechanism enables fusion models to dynamically prioritize reliable information and suppress misleading signals~\cite{feng2020deep,lou2023uncertainty,wang2023disthd}, which is crucial for enhancing the safety and dependability of AV perception, particularly in safety-critical edge cases~\cite{feng2020deep,xu2020squeezesegv3,chen2025hyperdimensional}.

\begin{figure}[!t]
\centering
\includegraphics[width=\linewidth]{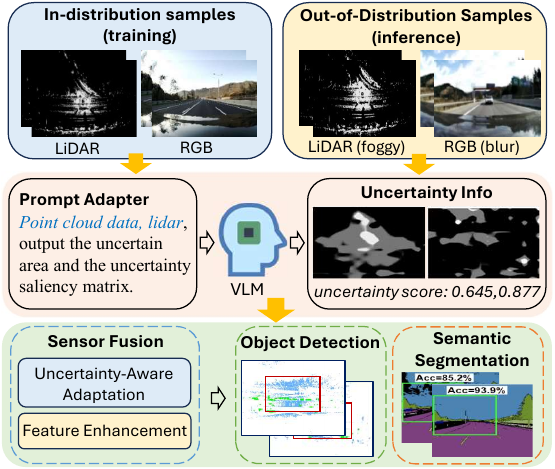}
\caption{ Overview of $\Design$. A VLM-guided UQ module provides pixel-level uncertainty maps to guide sensor fusion, leading to more robust object detection and semantic segmentation. 
 }
   \label{fig:motiv}
\end{figure}

Existing uncertainty-aware multi-modal sensor fusion typically adopt popular UQ techniques, such as Bayesian inference~\cite{lakshminarayanan2017simple,wang2023late}, ensemble methods~\cite{kwon2020uncertainty,wang2023domino}, and Monte Carlo dropout, to improve robustness~\cite{gal2016dropout}. 
However, a fundamental challenge remains: these data-driven approaches often struggle to generalize effectively in complex, out-of-distribution (OOD) scenarios, where conditions differ significantly from those seen during training~\cite{mukherjee2021decentralized,zhu2013variational,bijelic2020seeing}.
This limitation is especially critical for AVs, where encountering unseen conditions (e.g., severe weather or sudden sensor degradation) is inevitable in real-world deployments~\cite{wang2024rs2g,wang2025transformer}. 
A key contributing factor to this poor generalization is that most existing methods rely on coarse task-level or feature-level UQ, assigning a single uncertainty score to an entire modality or high-level representation. This results in an insufficient guidance signal. Consequently, they fail to capture the necessary fine-grained uncertainty variations at the pixel or voxel level, which are essential for dense prediction tasks like semantic segmentation and accurate 3D detection. 
Furthermore, these methods typically process each modality independently, often treating them as isolated data streams and lacking mechanisms to explicitly capture cross-modal dependencies~\cite{jung2023beyond,liang2023quantifying,wang2023hyperdetect,wang2024smore}. This deficiency prevents them from fully exploiting complementary sensor information, which is crucial when one sensor compensates for the weakness of another, leading to suboptimal representation learning and ineffective feature fusion.

To address these limitations, we propose $\Design$, a novel uncertainty-aware multi-modal sensor fusion framework that enhances the robustness and generalization of AV perception. Our approach is based on two key innovations. 
First, we leverage a Vision-Language Model (VLM) for fine-grained UQ. 
As shown in Figure \ref{fig:motiv}, we adapt a pre-trained VLM (Anomaly GPT~\cite{gu2023anomalyagpt}), originally developed for industrial anomaly detection, to the autonomous driving domain via LoRA fine-tuning~\cite{hu2022lora} on AV datasets. 
This adaptation allows the model to acquire not only domain-specific knowledge but also to reason over contextual relationships, which is vital for generalizing across various driving scenarios. 
{\textcolor{black}{Although VLM inference introduces additional computation, the response time of an autonomous system also depends on the computing hardware, inference implementation, and scheduling of perception, planning, and control. 
By combining local feature differences with scene context, the VLM provides uncertainty guidance that helps fusion retain useful information and suppress unreliable features under sensor degradation.}
During inference, the model first computes an uncertainty estimate by comparing the feature map of the test sample at each spatial location with representative training features. The VLM then refines this estimate by integrating local feature differences with global contextual reasoning, generating fine-grained, context-aware uncertainty heatmaps that highlight unreliable regions and guide the fusion process. 
Additionally, the VLM's strong transferability improves generalization to OOD scenarios by leveraging prior knowledge and robust anomaly localization abilities.
Second, to tackle the lack of cross-modal interaction, we further introduce a novel dynamic adaptation mechanism that explicitly captures cross-modal dependencies. This mechanism enables complementary sensor information (i.e., information where sensors compensate for each other's weaknesses) to be adaptively balanced during fusion, thereby achieving a more optimal and robust final representation. Our main contributions are summarized as follows:

\begin{itemize}[leftmargin=*]
    \item  We propose a $\Design$, a novel sensor fusion framework that incorporates a VLM-guided UQ module and a dynamic fusion mechanism to enhance multi-modal sensor fusion in AVs. 
    \textcolor{black}{Our approach achieves the highest mean performance among the compared sensor fusion methods on aiMotive for 3D object detection and on DeLiVER and MFNet for semantic segmentation.}
    
    \item \textcolor{black}{
    We introduce a VLM-guided UQ method that leverages prior knowledge and contextual understanding to refine the initial uncertainty estimates into context-aware, pixel-level uncertainty maps.
    }
    \textcolor{black}{It significantly improves generalization to complex, out-of-distribution driving scenarios, outperforming SOTA UQ techniques across all three datasets.}
    
    \item We design a novel fusion module that dynamically adjusts the contribution of each sensor with explicit modeling of cross-modal dependencies. This allows the model to exploit complementary information across modalities, 
    improving robustness across diverse real-world scenarios.

\end{itemize}

\section{Related Work}
\label{sec:rw}

\subsection{Uncertainty Quantification}
Various UQ techniques have been developed to enhance model robustness, including Bayesian neural networks that model weights as probability distributions~\cite{kwon2020uncertainty}, deep ensembles that estimate uncertainty by aggregating multiple predictions~\cite{lakshminarayanan2017simple}, and Monte Carlo (MC) dropout that introduces stochasticity during inference~\cite{gal2016dropout}. 
However, a critical limitation across these traditional methods is their failure to generalize effectively across environments, resulting in suboptimal performance in complex or OOD scenarios~\cite{gawlikowski2023survey}. 
Specifically, Bayesian models often suffer from poor scalability and rely on rigid probabilistic structures that fail to capture subtle contextual variations~\cite{kwon2020uncertainty, feng2020deep,sander2013bayesian}. Ensemble methods tend to overfit the training distribution, lacking the capacity to extrapolate to unseen conditions~\cite{xu2020squeezesegv3,brena2020choosing,rahaman2021uncertainty}. While MC dropout introduces randomness to approximate uncertainty, it lacks the necessary adaptability in dynamic environments~\cite {lou2023uncertainty,guan2023trustworthy}. In summary, these methods derive uncertainty largely from the statistical properties of the training data, making them inherently brittle when faced with distributional shifts. 
Recent advances in VLMs have demonstrated strong generalization capabilities across domains by leveraging large-scale pretraining~\cite{radford2021learning, li2023blip}. 
Unlike traditional models that rely solely on feature representations from a specific training distribution, VLMs incorporate extensive prior knowledge from vast and diverse datasets, enabling better adaptation to complex, OOD scenarios~\cite{zang2024overcoming,addepalli2024leveraging,chen2024practicaldg}. 
\textcolor{black}{However, recent studies identify overconfidence and hallucinations in VLMs~\cite{chen2025unveiling,guan2024hallusionbench}. In uncertainty-aware multi-modal sensor fusion, such errors may affect uncertainty estimates and sensor contributions, compromising the robustness of 3D object detection and semantic segmentation. DriveBench also highlights the potential of using VLMs to identify sensor degradation and improve perception reliability~\cite{xie2025drivebench}.}
VLMs' cross-modal semantic understanding and prior knowledge offers a powerful alternative to purely data-driven uncertainty estimation. However, translating the strong conceptual understanding of VLMs into concrete, pixel-level uncertainty maps for low-level perception in multi-modal UQ remains largely unexplored.

\subsection{Multi-modal Sensor Fusion in AVs}
Multi-modal sensor fusion is fundamental to AV perception, integrating features from diverse sensors to construct a comprehensive representation of the environment~\cite{chen2023futr3d,malawade2022hydrafusion}. Leveraging complementary strengths of different modalities, sensor fusion enhances performance in complex, dynamic scenarios where single-modality data may be unreliable~\cite{shekhar1986sensor,xu2018multi,huang2020multi}. In 3D detection, frameworks such as BEVFusion~\cite{liu2022bevfusion} and TransFusion~\cite{bai2022transfusion} align LiDAR point clouds and camera images via transformer-based architectures to capture both spatial depth and visual cues. 
For 2D segmentation, recent frameworks employ cross-modal attention~\cite{lu2023multi} and hierarchical fusion layers~\cite{wang2022multimodal} to adaptively weight sensor inputs~\cite{zhuang2021perception,meyer2019sensor}. 
Yet, despite their architectural sophistication, these approaches remain vulnerable to disturbances and OOD conditions such as adverse weather or sensor noise~\cite{bijelic2020seeing}. 
To mitigate these risks, a second category of fusion approaches has emerged, explicitly focusing on robustness across driving scenarios.
\textcolor{black}{
Earlier uncertainty-aware fusion methods address sensor degradation and previously unseen input conditions~\cite{tian2020uno,feng2020leveraging}.
}
Designs such as HydraFusion~\cite{malawade2022hydrafusion} and EcoFusion~\cite{malawade2022ecofusion} rely on coarse modality-level confidence, while more recent methods such as uncertainty-encoded fusion~\cite{lou2023uncertainty}, Cocoon~\cite{chococoon}, SAMFusion~\cite{palladin2024samfusion}, CAFuser~\cite{brodermann2025cafuser}, and ContextualFusion~\cite{sural2024contextualfusion} introduce condition- or uncertainty-aware weighting to adapt sensor contributions under adverse conditions. 
\textcolor{black}{
HyperDUM~\cite{chen2025hyperdimensional} further considers spatial- and channel-wise uncertainty, while VLC Fusion~\cite{taparia2025vlc} uses VLM-based environmental context to dynamically adjust sensor contributions.
}
\textcolor{black}{
However, using VLM contextual reasoning to refine initial uncertainty estimates into context-aware, pixel-level uncertainty maps for subsequent fusion remains largely unexplored.
}
Furthermore, few methods explicitly model the uncertainty-driven interaction between modalities, which limits their ability to dynamically and precisely leverage complementary information. 
\textcolor{black}{
$\Design$ addresses these limitations by using the VLM to refine the initial uncertainty estimates into context-aware, pixel-level uncertainty maps, which are then used to suppress unreliable regions and guide the subsequent modality dynamic adaptation.
}

\begin{figure*}[t] 
\centering
    \includegraphics[width=1\textwidth]{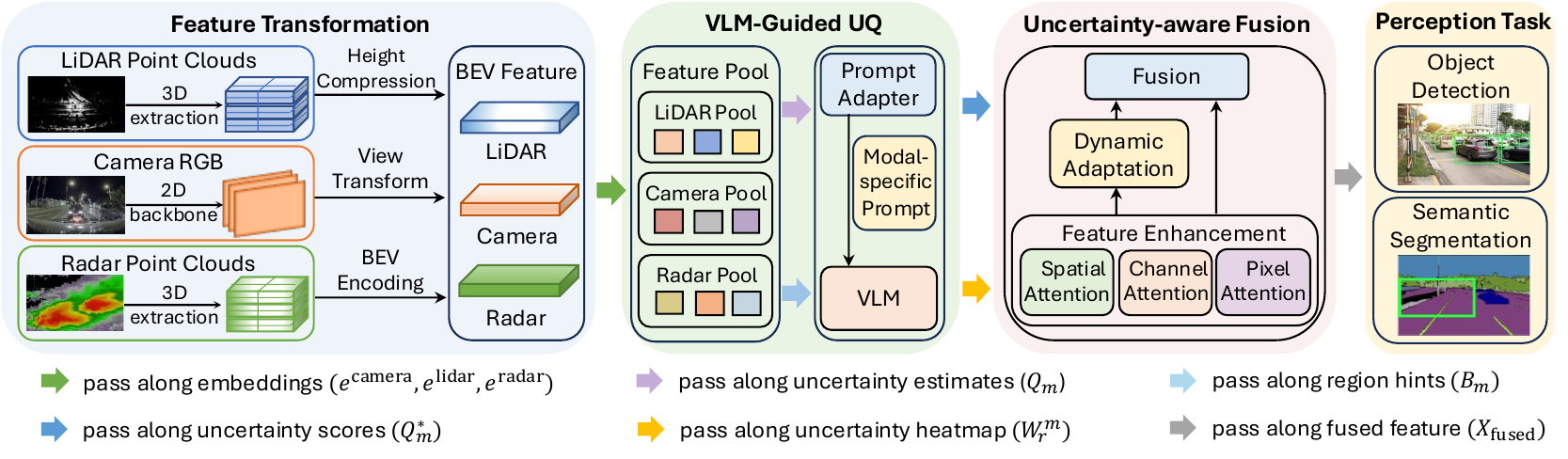}
 \vspace{-6mm}
    \caption{3D architecture of our uncertainty-aware multi-modal sensor fusion framework for AV perception tasks. The framework comprises three states: feature transformation into the BEV space; a VLM-guided UQ module that generates fine-grained uncertainty heatmaps using modality-specific prompts; and an uncertainty-aware fusion module that leverages a dynamic adaptation mechanism to produce fused features for object detection and semantic segmentation.}
    \label{fig:workflow}
\vspace{-1mm}
\end{figure*}
\section{Methodology}\label{sec:method}

\subsection{Problem Formulation}
We propose $\Design$, as demonstrated in Figure \ref{fig:workflow}, a novel uncertainty-aware sensor fusion framework structured into two main stages: VLM-guided UQ and uncertainty-aware fusion. 
$\Design$ first projects features captured by different modalities into a unified shared feature space, typically a Bird's-Eye-View (BEV) representation, to ensure feature alignment. A prompt adapter is then fine-tuned for a VLM (AnomalyGPT~\cite{gu2023anomalyagpt}) using these feature representations to allow the model to acquire essential prior knowledge specific to autonomous driving. 
During training, we construct a feature pool for each modality, storing features of representative training samples as a reference for uncertainty estimation. 
Subsequently, during inference, we compute an initial uncertainty estimate by comparing the test sample’s feature map with these representative training samples in the feature pool. 
The VLM then refines these estimates through global contextual reasoning, leveraging its deep semantic understanding to generate a highly informative uncertainty heatmap that accurately localizes unreliable regions at a fine-grained spatial level. 
Notably, our model-agnostic UQ module can be integrated into existing fusion frameworks and extended to additional modalities via finetuning with corresponding data. 
In the uncertainty-aware fusion stage, $\Design$ improves representation learning through spatial, channel, and pixel attention to emphasize critical regions, highlight informative features, and adjust the modality-specific importance. 
We further introduce a novel dynamic adaptation mechanism that explicitly models cross-modal dependencies, enabling modalities to reinforce reliable signals and compensate for uncertain features. Finally, $\Design$ adaptively fuses features based on the fine-grained uncertainty information provided by the VLM-guided UQ, thereby effectively mitigating error propagation and significantly enhancing robustness in downstream perception tasks.

\subsection{Multi-Modal Feature Extraction}
We employ modality-specific backbones to extract feature representations from LiDAR point clouds, radar point clouds, and multi-view camera images. 
These heterogeneous features are then projected into a shared feature space using learnable projection functions to ensure cross-modal alignment.

\subsubsection{LiDAR Features:}  We use VoxelNet~\cite{zhou2018voxelnet} with SparseConvNet~\cite{wang2023large} to extract LiDAR features, encoding voxel-wise geometric properties into feature vectors that capture 3D spatial structures. The extracted feature for the $i^\text{th}$ LiDAR sample is denoted as $ \mathbf{f}_l^{(i)} \in \mathbb{R}^{C_l \times D_l \times H_l \times W_l} $, where $ C_l $ is the number of feature channels, and $ D_l $, $ H_l $, and $ W_l $ denote the depth, height, and width of the 3D feature map, respectively. 
We then apply height compression to obtain a 2D bird's eye view (BEV)  representation $\mathbf{e}_l^{(i)} \in \mathbb{R}^{C \times H \times W}$, where $C$ is the number of feature channels, $H$ and $W$ are the height and width of the shared feature space, respectively. This transformation preserves key geometric information while ensuring alignment with the shared feature space. 

\subsubsection{Camera Features:} 
We use ResNet-50~\cite{koonce2021resnet} to extract high-level image (RGB) features from multi-view camera images. The extracted feature for the $i^\text{th}$ sample from the $j^\text{th}$ view is denoted as $\mathbf{f}_c^{(i,j)} \in \mathbb{R}^{C_c \times H_c \times W_c}$, where $C_c$ is the number of feature channels, and $H_c$ and $W_c$ are the height and width of the feature map, respectively.  
To ensure spatial alignment with other modalities, we apply view transformation to project image features into the BEV space, followed by a $1 \times 1$ convolution to map $C_c$ channels to $C$ channels.  The final camera representation is $\mathbf{e}_c^{(i,j)} \in \mathbb{R}^{C \times H \times W}$, where $C$ is the number of feature channels, $H$ and $W$ are the height and width of the shared feature space, respectively.

\subsubsection{Radar Features:} 
We use PointPillars~\cite{lang2019pointpillars} to extract radar features by voxelizing radar point clouds into a sparse pillar-based BEV representation. Each pillar aggregates radar reflections, capturing both spatial distribution and Doppler velocity information. 
The extracted representation for the $i^\text{th}$ radar sample is denoted as $ \mathbf{f}_r^{(i)} \in \mathbb{R}^{C_r \times H_r \times W_r}$, where $C_r$ is the number of feature channels, and $H_r, W_r$ are the height, and width of the BEV feature map. 
The radar features are then processed using a SparseCNN~\cite{graham2017submanifold} to extract global spatial features while preserving motion information. Finally, a $1 \times 1$ convolution is applied to ensure feature compatibility, producing the final radar feature representation as $\mathbf{e}_r^{(i)} \in \mathbb{R}^{C \times H \times W}$, where $C, H, W$ are the number of feature channels, height, and width of the shared feature space.

\subsection{Feature Pool}\label{subsec:pool}

\subsubsection{Core Sampling} 
Each training sample comprises modality-specific inputs from camera, radar and LiDAR with feature representations $\mathbf{e}_i^\text{camera},  \mathbf{e}_i^\text{LiDAR}, \mathbf{e}_i^\text{radar}
\in \mathbb{R}^{C \times H \times W} (1\leq i\leq N)$, where $N$ is the number of training samples, and $C, H, W$ denote the number of feature channels, height, and width of the shared feature space, respectively. 
Note that these feature representations retain their spatial structure rather than being flattened vectors. 
To reduce memory and computational cost, we apply core sampling~\cite{roth2022towards} to construct feature pools $\mathcal P_\text{camera},\mathcal P_\text{LiDAR}, \mathcal P_\text{radar} \in \mathbb{R}^{N'\times C \times H \times W} $, selecting $N'$ ($N' \ll N$) representative features for each modality. 
Denoting the full set of feature representations of each modality as $\mathcal E_m \in \mathbb{R}^{N \times C \times H \times W}$, where $m\in\{\text{camera, LiDAR, radar}\}$, we define the closest representative $\mathbf p\in\mathcal P_m$ for each $\mathbf{e} \in \mathcal E_m$ using the Euclidean distance: 
\begin{equation}
d(\mathbf{e}, \mathcal P_m)= \min_{\mathbf p \in \mathcal{P}_m} \sum_{i=1}^{H} \sum_{j=1}^{W}  \| \mathbf e_{:, i, j} - \mathbf p_{:, i, j} \|_2.
\label{eq:nearest_neighbor_distance}
\end{equation}  
We iteratively add the feature representation with the largest distance $d$ to the pool until it reaches $N'$ samples. 
By minimizing the worst-case distance between features in $\mathcal E_m$  and their nearest representatives in $\mathcal P_m$, the feature pool efficiently preserves the diversity of training samples. 
\textcolor{black}{
The sampling distance in Eq.~\ref{eq:nearest_neighbor_distance} is computed over the full feature representation, while the selected feature representations retain their spatial structure to support subsequent uncertainty estimation at each spatial location.
}
At inference, we estimate uncertainty at each spatial location in the test sample’s feature representation by comparing it to the embeddings at the same location in the representations stored in the feature pool of the corresponding modality.

\subsubsection{Uncertainty Estimation} 
At inference, we compute the uncertainty estimates $Q_m \in \mathbb{R}^{H \times W}$ for a test sample $ \mathbf{v} \in \mathbb{R}^{C \times H \times W}$ of modality $m$. 
Each element $Q_{ij}^m (0\leq i\leq H, 0\leq j\leq W)$ in $Q_m,$ represents the minimum Euclidean distance between the feature vector $\mathbf{v}_{ij} \in \mathbb{R}^{C}$ at spatial location $(i,j)$ in the feature representation of the test sample and  its closest feature vector $\mathbf{p}_{ij} \in \mathbb{R}^{C}$ at the same location of all the representative features in the feature pool, i.e.,  
\begin{equation}
Q^m_{ij} = \min_{\mathbf p \in \mathcal P_m} \left\| \mathbf{v}_{ij} - \mathbf{p}_{ij} \right\|_2.
\label{eq:confidence_score}
\end{equation}
\textcolor{black}{
A higher $Q^m_{ij}$ value indicates a larger deviation from the representative training features at location $(i,j)$.
Since such a deviation may result from either sensor degradation or a valid unseen scene, $Q_m$ is used as an initial uncertainty estimate and is further refined by the VLM.
}
Additionally, we introduce region hints $B_m\in \mathbb{R}^{H\times W}$ by thresholding $Q_m$, highlighting high-uncertainty regions in the feature representation. 
The uncertainty estimates and region hints are then passed to the VLM, integrating local uncertainty estimates with contextual reasoning.

\subsection{VLM-Guided Uncertainty Quantification}
We fine-tune AnomalyGPT~\cite{gu2023anomalyagpt} using LoRA~\cite{hu2022lora} in an unsupervised manner to align visual uncertainty cues (i.e., localization map $M$) with language-based reasoning (i.e., uncertainty map $A_\text{text}$). AnomalyGPT generates uncertainty heatmaps that localize abnormal regions based on both low-level details and high-level semantic context. We use AnomalyGPT for its dual-stream design and strong localization, but our method can be generalized to any VLM supporting vision-text alignment and region-level output. 
AnomalyGPT is designed to produce uncertainty heatmaps that localize abnormal regions in input images. It employs convolutional front-end layers to capture low-level details, followed by attention mechanisms for high-level semantic reasoning.

\subsubsection{Visual Stream}
The input feature representations to the VLM are, $\mathbf{e}_i^\text{camera}, \mathbf{e}_i^\text{LiDAR}, \mathbf{e}_i^\text{radar}\in \mathbb{R}^{C \times H \times W}$, where $i$ indexes the sample, $C$ is the number of channels, and $H$ and $W$ are the height and width of the shared feature space, respectively. 
These features are processed by a frozen ImageBind encoder~\cite{girdhar2023imagebind}, which extracts multi-scale patch-level features from hierarchical layers, generating feature maps $F_j^{\text{patch}}$, where $j$ indexes the hierarchical layer depth. 
These features are then passed through a lightweight transformer-based decoder~\cite{gu2023anomalyagpt} that aligns visual features with learned textual semantics. The decoder performs visual-textual feature matching and outputs a fine-grained uncertainty localization map $M \in \mathbb{R}^{H \times W}$. 
Specifically, it aligns the intermediate patch-level features $F_j^{\text{patch}}$ with learned textual features $ F_{\text{text}}$, which capture semantics of normal and abnormal scenarios, enabling fine-grained uncertainty localization and quantification across diverse environments. 
The decoder computes similarity scores between patch-level features and textual
features, then applies up-sampling to generate a high-resolution localization map $M\in \mathbb{R}^{H \times W}$, indicating uncertainty regions in the feature. 

\subsubsection{Textual Stream}
Powered by Vicuna-7B~\cite{chiang2023vicuna}, the textual stream interprets uncertainty via attention-based reasoning using four inputs:  
(i) a localization map $M$ from the visual stream, 
(ii) a user query $q_{\text{user}}$ (e.g., ``Describe the detected uncertainty in this image''), 
(iii) a structured prompt $E_{\text{prompt}}$ from the prompt adapter (details below), and 
(iv) feature representations.
These inputs are processed by Vicuna-7B, generating an attention-based uncertainty map $A_\text{text}$ from cross-modal attention scores over $E_{\text{region}}$. This map highlights uncertain regions, and a context-aware pixel-level uncertainty scores $Q_m^* \in \mathbb{R}^{H\times W}$ is obtained by interpolating attention scores to match the resolution of the visual stream.

\begin{itemize}[leftmargin=*]
\item{\textbf{Prompt Adapter: }}
To bridge the vision and textual streams, we introduce a prompt adapter that converts uncertainty estimates $Q_m$,  region hints $B_m$  $(m\in\{\text{camera, LiDAR, radar}\})$, 
and localization maps $M$ into structured prompts. 
The structured prompt is formed by concatenating a base prompt $E_{\text{base}}$, an uncertainty embedding $E_{\text{uncertain}}$, and refined region hints $E_{\text{region}}$. 
Specifically, the base prompt $E_{\text{base}}$ provides uncertainty detection objective and modality specification  (e.g., camera, LiDAR, radar).
The uncertainty embedding $E_{\text{uncertain}}$ is derived from the uncertainty estimates $Q_m$ and compressed with a CNN~\cite{lecun1998gradient} to ensure compatibility. 
The refined region hints $E_{\text{region}}$ is constructed from the localization map $M$ and the 
region hints $B_m$  with positional encoding to emphasize high-uncertainty regions.
The final structured prompt is defined as
$E_{\text{prompt}} = [E_{\text{base}}, E_{\text{uncertainty}}, E_{\text{region}}]$. 
\end{itemize}
\subsubsection{Integration of Visual and Textual Streams}
We integrate the outputs
from both the visual and textual streams to produce the final uncertainty heatmap $W_r^m$ that  precisely localizes uncertainty areas in the original images: 
\begin{equation}
W_r^m = \lambda M + \gamma A_{\text{text}}
\label{eq:w_r^m}
\end{equation}
where $M$ is the visually derived localization map,  $A_{\text{text}}$ is the contextually derived uncertainty map, and $\lambda$ and $\gamma$ are hyperparameters that control the relative contributions of the visual and textual information, respectively. 
\textcolor{black}{
To reduce the effect of unreliable textual reasoning, $A_{\text{text}}$ is combined with the visually derived localization map $M$ rather than being used independently, while $\gamma$ controls the contribution of the textual stream to $W_r^m$.
}
This final heatmap $W_r^m$ provides the refined, context-aware guidance signal used in the subsequent uncertainty-aware fusion stage.

\subsection{Uncertainty-aware Sensor Fusion}

We propose a novel uncertainty-aware fusion module that adaptively fuses features from each sensor based on the UQ information provided by the VLM to enhance robustness.
We first use the uncertainty heatmap $W_r^m$ as a mask to suppress unreliable regions, generating a refined representation  
\begin{equation}
X'_i = X_i \odot (1 - W_r^m)
\end{equation}
where $\odot$ denotes element-wise multiplication and $X_i$ is the initial feature representation of modality $m$. We reduce uncertainty propagation by assigning smaller weights to high-uncertainty regions in $W_r^m $. 
\subsubsection{Feature Enhancement} \label{subsubsec:feature_enhance}

We employ a combination of spatial, channel, and pixel-level attention mechanisms to capture comprehensive contextual information.

\begin{itemize}[leftmargin=*]

\item{\textbf{Spatial Attention:}} 
The spatial attention mechanism captures spatial dependencies by generating a weight map $S_i$ for each modality’s feature map $X_i'$, highlighting high-confidence (low-uncertainty) regions.
We apply global average pooling $X'_{\text{avg}}$ and max pooling $X'_{\text{max}} $ across the channels of $X'_i$, concatenate the descriptors, and process them through a convolutional layer to refine the attention weights:
\begin{equation}
S_i = \sigma(\text{Conv2D}(\text{Concat}(X'_{\text{avg}}, X'_{\text{max}}))),
\end{equation}
where $\sigma$ denotes the sigmoid activation function.

\item{\textbf{Channel Attention:}}
The channel attention mechanism enhances informative channels by dynamically adjusting their importance. We apply global average pooling on $X'_i$ to obtain a channel descriptor $T_i$, which is then passed through a two-layer multi-layer perceptron to generate channel-wise attention weights $T'_i$. 
These weights emphasize significant channels while suppressing less useful ones. The enhanced feature representation is computed as:
\begin{equation}
X''_i = X'_i \odot S_i \odot T'_i,
\end{equation} 
where \( \odot \) denotes element-wise multiplication. The resulting feature map \( X''_i \) effectively integrates spatial- and channel-wise significant features, emphasizing relevant information while suppressing noisy or unreliable regions. 

\item{\textbf{Pixel Attention:}} 
The pixel attention mechanism facilitates cross-modal fusion by adaptively weighting each modality at the pixel level. 
We concatenate features from different modalities and process them through a convolutional layer to produce a pixel-wise attention map  $G_i$:
\begin{equation}
G_i = \sigma(\text{Conv2D}(\text{Concat}(X''_i, X''_{\text{o}})))
\label{eq:pixel_wise_weight}
\end{equation}
where $X''_{\text{o}}$ denotes the feature map of another modality that has also undergone spatial and channel attention.
The final fused representation is computed as
\begin{equation}
X_{\text{final}} = G_i \odot X''_{\text{o}} + (1 - G_i) \odot X''_i
\label{eq:fused_final}
\end{equation}
where $\odot$ denotes element-wise multiplication. This mechanism adaptively adjusts the contribution of each modality at the pixel level, emphasizing high-confidence regions while suppressing unreliable information. We extend Eq. (\ref{eq:pixel_wise_weight})(\ref{eq:fused_final}) to three modalities fusion by sequentially applying the same fusion step twice: first fusing two modalities (e.g., LiDAR + camera), then fusing the result with the third (e.g., radar). 
\end{itemize}

\subsubsection{Modality Dynamic Adaptation} \label{subsubsec:dynamic_adaptation}
We introduce a novel dynamic adaptation mechanism that adaptively adjusts the weight of each modality in feature fusion based on the uncertainty score $Q'_m$, a normalized scalar in $[0,1]$ obtained by aggregating pixel-wise uncertainty scores $Q_m^*$. 
\textcolor{black}{
While $W_r^m$ provides the refined uncertainty heatmap used to suppress unreliable regions, $Q'_m$ summarizes the pixel-wise uncertainty scores as a scalar to adjust the contribution of each modality.
}
This mechanism ensures that more reliable modalities receive higher weights and contribute more to the fused representation.
The weighting function achieves this by incorporating both uncertainty scores and inter-modality variance. 
\textcolor{black}{Specifically, the weight $w_m$ for each modality $m\in\{\text{camera, LiDAR, radar}\}$ is defined as:
\begin{equation}
w_m =
\frac{
\exp(-\nu \cdot Q'_m)
+\gamma \cdot \frac{1}{K}
\sum_{k=1}^{K}(Q'_k-\bar Q)^2
}{
\sum_{k=1}^{K}
\left(
\exp(-\nu \cdot Q'_k)
+\gamma \cdot \frac{1}{K}
\sum_{j=1}^{K}(Q'_j-\bar Q)^2
\right)
}.
\label{eq:weight_1}
\end{equation}
where $K$ is the total number of modalities,} $\bar{Q}$ is the mean uncertainty score across all the modalities, $\nu$ controls sensitivity to the uncertainty score, and  $\gamma$ adjusts sensitivity for inter-modality variance.
By dynamically weighting modalities based on their uncertainty and inter-modality variance, this adaptive weighting enhances feature fusion by explicitly incorporating cross-modal dependencies. 
The final fused feature $X_{\text{fused}}$ is then computed 
as a weighted sum of each modality's feature representation $X_{\text{final}}^{m}$: 
\textcolor{black}{\begin{equation}
    X_{\mathrm{fused}}
=
\sum_m w_m X_{\mathrm{final}}^m.
\end{equation}}
 ${X_{\text{fused}}}$ is then passed to a task head to perform downstream perception tasks. The task-specific loss is denoted as $L_\text{baseline}$. 

\subsubsection{Modality Consistency Loss}
We introduce a UQ loss and a modality consistency loss to reduce inconsistencies across modalities and stabilize representations in uncertain regions. 
\begin{itemize}[leftmargin=*]
\item{\textbf{UQ Loss:}} The UQ loss $L_{\text{uncertainty}}$ penalizes the differences between feature representations in identified high-uncertainty regions $\mathbf{e}_r^m$ and their closest representative  features $\mathbf{e}_l^m$ from the feature pool $\mathcal P_m$, where $m\in\{\text{camera, LiDAR, radar}\}$. 
By minimizing this loss, the model stabilizes feature representations in uncertain areas and prevents significant deviation from the representative features in the feature pool. 
This promotes feature consistency and reduces uncertainty propagation, improving the robustness of downstream tasks. 
Mathematically,
\begin{equation}
L_{\text{uncertainty}} = \sum_m \sum_{r} \mathbb{I}(W_r^m > \tau) \cdot \| \mathbf{e}_r^m - \mathbf{e}_l^m \|_2,
\label{eq:q_loss}
\end{equation}
where $\mathbb{I}$ is an indicator function that activates for each pixel $(i,j)$ in the heatmap $W_r^m$ that exceeds a threshold $\tau$.

\item{\textbf{Modality Consistency Loss:}} The modality consistency loss $L_{\text{consistency}}$ enforces feature alignment across modalities in high confidence (low-uncertainty) regions, ensuring coherent multi-modal fusion. Mathematically, 
\begin{equation}
L_{\text{consistency}} = \sum_{m_1 \neq m_2} \sum_{r \in \mathbb{I}(W_r^m \leq \tau)} \| \mathbf{e}_r^{m_1} - \mathbf{e}_r^{m_2} \|_2,
\label{eq:r_loss}
\end{equation}
where $m_1$ and $m_2$ denote different modalities, $\tau$ denote a predefined threshold, and $r$ denotes the uncertain regions. 
The total loss function is then defined as:
\begin{equation}
L_{\text{total}} = L_{\text{baseline}} + \alpha L_{\text{uncertainty}} + \beta L_{\text{consistency}},
\label{eq:total_loss}
\end{equation}
where 
$\alpha$ and $\beta$ are coefficients adjusting the importance of UQ loss and modality consistency loss, respectively. 
\end{itemize}
\section{Experimental Evaluation}\label{sec:result}

\subsection{Experimental Setup}

\begin{figure}[!t]
\vspace{-2mm}
\centering
\includegraphics[width=\linewidth]{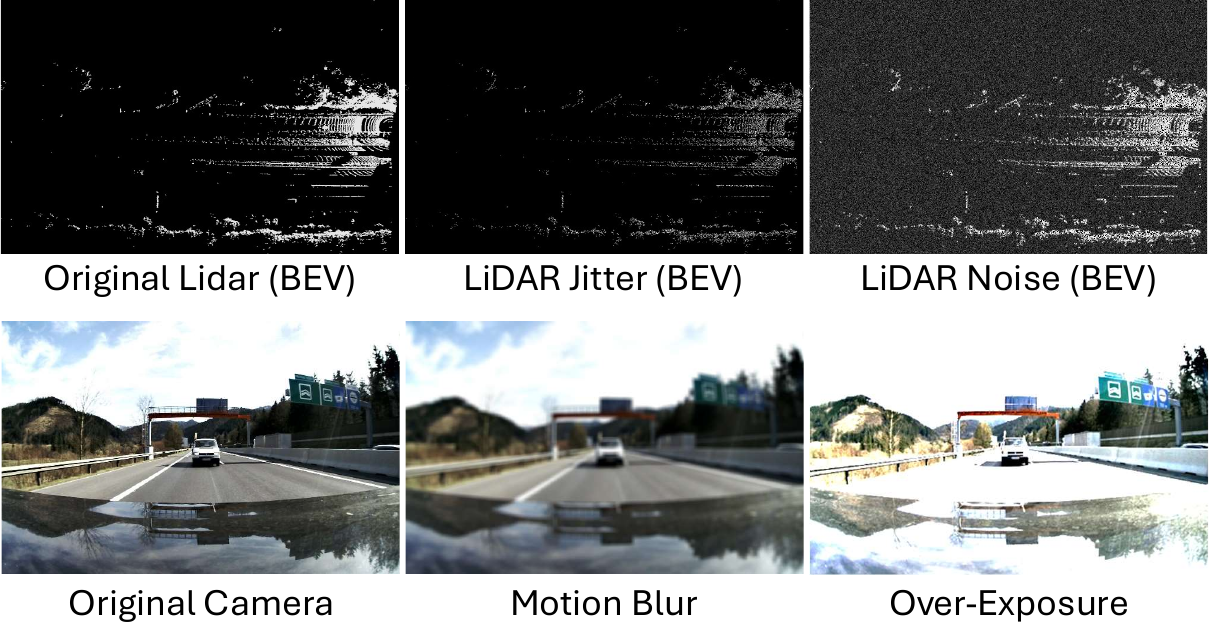}
\caption{Original and Simulated 
    Out-of-Distribution Samples: The figure displays  LiDAR data perturbed by jitter and noise, and camera images degraded by motion blur and over-exposure. These samples, alongside the original inputs, are utilized for robust evaluation under OOD conditions. }
   \label{fig:simulate}
 \vspace{-3mm}
\end{figure}

\noindent\textbf{Datasets:} We evaluate $\Design$ under various adverse conditions using three public autonomous driving datasets. 
\begin{itemize}
    \item \textbf{aiMotive}~\cite{matuszka2022aimotive}: 
    The dataset consists of 26,583 annotated frames of camera and LiDAR data for 3D object detection, covering over 425,000 objects across 14 classes. It spans diverse driving conditions (highway, night, rain, and urban) and is split into 21,402 training frames and 5,181 validation frames. We simulated 5181 out-of-distribution (OOD) scenarios for inference by applying motion blur (MB), over-exposure (OE), under-exposure (UE), and LiDAR noise (LN).
    \item \textbf{DeliVER}~\cite{zhang2023delivering}: 
   The dataset consists of 47,310 samples, including 1406 OOD samples, with six views and four sensor modalities (cameras, LiDAR, event, and depth sensors) for semantic segmentation under challenging conditions (cloudy, foggy, night, rainy). 
   Our OOD simulation includes MB, OE, UE, LN, and event low-resolution.

    \item \textbf{MFNet}~\cite{ha2017mfnet}: 
    This dataset consists of 1,569 paired visual (RGB) and thermal infrared images, including 820 daytime and 749 nighttime samples.
    We simulated 392 OOD samples for inference. 
\end{itemize}
\noindent\textbf{Evaluation Metrics: }Following SOTA research, we evaluate $\Design$ on the aiMotive dataset using all-point/11-point interpolation average precision (AP) metrics for object detection, and on DeLiVER and MFNet using mean Intersection over Union (mIoU) for semantic segmentation.

\begin{table*}[htbp]
\centering
\footnotesize 
\addtolength{\tabcolsep}{-0.35em}{\begin{tabular}{l c c c c c c c }
\toprule
\multirow{2}{4em}{\textbf{Scenarios}} & \multicolumn{5}{c}{\textbf{Sensor Fusion Method}} & \textbf{Baseline} & \multicolumn{1}{c}{\textbf{Ours}} \\
\cmidrule(lr){2-8} 
& 
FocalFormer3D~\cite{focalformer3d} & MVP~\cite{yin2021multimodal} & BEVFusion~\cite{liu2022bevfusion} & CMT~\cite{yan2023cross} & MSMDFusion~\cite{Jiao_2023_CVPR} & aiMotive~\cite{matuszka2023aimotive} & $\Design$ \\
\midrule
\textbf{Highway} & 69.94/67.72 & 68.74/66.60 & \colorbox{palered}{76.89/74.64} & 73.50/70.98 & 75.64/73.56 & 75.67/73.50 & \colorbox{lavender}{76.27/74.09} \scriptsize{{\textcolor{blue}{(+0.60/+0.59)}}} \\
\textbf{Urban} & 56.96/56.16 & 59.71/58.49 & 58.37/57.70 & 60.18/59.01 & \colorbox{lavender}{61.77/60.49} & 61.10/60.01 & \colorbox{palered}{63.29/62.23} \scriptsize{{\textcolor{blue}{(+2.19/+2.22)}}} \\
\textbf{Night} & 66.42/64.75 & 70.12/68.15 & 73.08/71.54 & 71.22/69.38 & \colorbox{lavender}{74.21/72.08} & 73.44/71.21 & \colorbox{palered}{75.96/72.56} \scriptsize{{\textcolor{blue}{(+2.52/+1.35)}}} \\
\textbf{Rain} & 40.42/40.19 & 42.85/41.75 & \colorbox{lavender}{44.92/43.87} & 43.01/41.75 & \colorbox{palered}{45.01/43.92} & 41.18/40.21 & 44.17/43.14 \scriptsize{{\textcolor{blue}{(+2.99/+2.93)}}} \\
\textbf{Mean} & 61.21/59.95 & 63.36/61.87 & 67.73/\colorbox{lavender}{66.06} & 65.10/63.80 & \colorbox{lavender}{67.76}/65.96 & 65.85/64.46 & \colorbox{palered}{68.37/66.76} \scriptsize{{\textcolor{blue}{(+2.52/+2.30)}}} \\
\bottomrule
\end{tabular}}
\vspace{4mm}
\caption{Multi-sensor fusion performance on the aiMotive dataset. Metrics are reported as all-point AP/11-point interpolation AP (in \%). The best (second-best) results are highlighted in \colorbox{palered}{pink} (\colorbox{lavender}{purple}). Improvements over the baseline are shown in {\textcolor{blue}{blue}}.}
\vspace{-2mm}
\label{table:fusion1}
\end{table*}

\begin{table*}[htbp]
\centering
\footnotesize 
\addtolength{\tabcolsep}{0.2em}{\begin{tabular}{l c c c c c c }
\toprule
\multirow{2}{4em}{\textbf{Scenarios}} & \multicolumn{4}{c}{\textbf{Sensor Fusion Method}} & \textbf{Baseline} & \textbf{Ours} \\
\cmidrule(lr){2-7} 
& StitchFusion~\cite{li2024stitchfusion} & CMX~\cite{zhang2023cmx} & TokenFusion~\cite{wang2022multimodal} & HRFuser~\cite{broedermann2023hrfuser} & CMNeXt~\cite{zhang2023delivering} & $\Design$ \\
\midrule
\textbf{Cloudy} & \colorbox{lavender}{50.63}/55.89 & 48.96/56.00 & 43.10/49.19 & 37.56/39.69 & {50.48/\colorbox{lavender}{56.61}} & \colorbox{palered}{53.93/58.25} \scriptsize{\textcolor{blue}{(+3.45/+1.64)}} \\
\textbf{Foggy}  & \colorbox{lavender}{50.65}/55.12 & 49.93/54.24 & 43.57/48.88 & 35.10/40.53 & {50.58/\colorbox{lavender}{55.13}} & \colorbox{palered}{54.27/57.94}\scriptsize{\textcolor{blue}{(+3.66/+2.81)}} \\
\textbf{Night}  & 44.65/\colorbox{lavender}{49.21} & 42.75/47.47 & 40.13/41.04 & 30.99/33.49 & \colorbox{lavender}{44.95}/{48.02} & \colorbox{palered}{47.19/49.72} \scriptsize{\textcolor{blue}{(+2.24/+1.70)}} \\
\textbf{Rainy}  & \colorbox{lavender}{52.09/56.33} & 51.39/55.68 & 45.69/49.02 & 39.40/41.12 & 51.34/55.05 & \colorbox{palered}{55.52/58.03} \scriptsize{\textcolor{blue}{(+4.18/+2.98)}} \\
\textbf{Sunny}  & \colorbox{lavender}{50.30}/{57.35} & 49.15/\colorbox{lavender}{57.87} & 45.14/51.08 & 38.41/42.17 & 49.24/57.14 & \colorbox{palered}{52.25/58.01} \scriptsize{\textcolor{blue}{(+3.01/+0.87)}} \\
\textbf{Mean}   & \colorbox{lavender}{49.67/54.88} & 48.44/54.25 & 43.53/47.84 & 36.29/39.40 & {49.32/54.39} & \colorbox{palered}{52.63/56.79 }\scriptsize{\textcolor{blue}{(+3.31/+2.40)}} \\
\bottomrule
\end{tabular}}
\vspace{4mm}
\caption{Multi-Modal Sensor Fusion for RGB-LiDAR on DeLiVER test set (out-of-distribution data)/validation set (in-distribution data) under Adverse Weather Conditions (Metrics: mIoU). The best (second-best) results are highlighted in \colorbox{palered}{pink} (\colorbox{lavender}{purple}). Improvements over the baseline are indicated in {\textcolor{blue}{blue}}.}
\label{table:fusion2}
\vspace{-2mm}
\end{table*}

\begin{table}[htbp]
\centering
\footnotesize
\addtolength{\tabcolsep}{-0.3em}{
    \begin{tabular}{l c c c |c}
\toprule
\textbf{Method}
 &\textbf{Modality} &\begin{tabular}[c]{@{}c@{}}\textbf{UQ}\\ \textbf{Method}\end{tabular} & \begin{tabular}[c]{@{}c@{}}\textbf{mIoU} \\ \textbf{(\%)} \end{tabular} & \begin{tabular}[c]{@{}c@{}}\textbf{Inference} \\\textbf{Latency} \end{tabular}\\
\midrule
SwinT~\cite{liu2021swin}        & RGB &\xmark & 40.23 & $-$\\
SegFormer~\cite{xie2021segformer}   & RGB &\xmark & 43.21 & $-$\\
\midrule
ACNet~\cite{hu2019acnet}        & RGB-Thermal &\xmark &44.80 & $-$\\
FuseSeg~\cite{sun2020fuseseg}   & RGB-Thermal &\xmark &51.44 & $-$\\
FEANet~\cite{deng2021feanet}   & RGB-Thermal &\xmark &53.49 & $-$\\
SegMiF~\cite{liu2023segmif}      & RGB-Thermal &\xmark &53.98 & $-$\\
CMX (B4)~\cite{zhang2023cmx}     & RGB-Thermal &\xmark &57.14 & $-$\\
CMNeXt~\cite{zhang2023delivering}      & RGB-Thermal &\xmark &{58.47} & $-$\\
\midrule
CMNeXt~\cite{zhang2023delivering}     & RGB-Thermal &{InfMCD~\cite{mi2022training}} &  58.62 & \colorbox{lavender}{0.56s}\\
CMNeXt~\cite{zhang2023delivering}     & RGB-Thermal &{InfNoise~\cite{mi2022training}} & \colorbox{lavender}{58.86} & 0.59s\\
CMNeXt~\cite{zhang2023delivering}     & RGB-Thermal &{LDU~\cite{franchi2022latent}} & 58.58 & \colorbox{palered}{0.50s}\\
CMNeXt~\cite{zhang2023delivering}     & RGB-Thermal &{$\Design$(ours)} & \colorbox{palered}{59.78} & 0.66s\\
\bottomrule
\end{tabular}}
\vspace{3mm}
\caption{Multi-sensor fusion performance on MFNet \cite{ha2017mfnet}. The best (second-best) results are highlighted in \colorbox{palered}{pink}(\colorbox{lavender}{purple}).}
\label{table:fusion3}
\vspace{-8mm}
\end{table}

\begin{table*}[htbp]
\centering
\footnotesize
\begin{tabular}{l c c c c c | c}
\toprule
\textbf{UQ Method} & \textbf{Motion-Blur} & \textbf{Over-Exposure} & \textbf{Under-Exposure} & \textbf{LiDar-Noise} & \textbf{Mean} & \textbf{Inference Latency}\\
\midrule
PostNet~\cite{charpentier2020posterior} & 60.42/58.90 & 63.12/59.99 & 60.20/59.25 & 64.13/60.60 & 61.97/59.69 &$-$\\
InfMCD~\cite{mi2022training}  & 62.00/61.99 & 64.47/64.28 & \colorbox{lavender}{63.25/63.49} & 64.27/63.81 & 63.50/\colorbox{lavender}{63.39} & \colorbox{lavender}{0.60s}\\
InfNoise~\cite{mi2022training}  & \colorbox{lavender}{62.61/62.61} & 64.94/\colorbox{lavender}{64.72} & 63.18/60.07 & 65.19/\colorbox{lavender}{64.99} & \colorbox{lavender}{63.98}/{63.10} & 0.62s\\
LDU~\cite{franchi2022latent}  & 62.06/62.22 & \colorbox{lavender}{65.07}/\colorbox{palered}{64.85} & 62.87/60.02 & \colorbox{lavender}{65.47}/\colorbox{palered}{65.10} & 63.87/63.05 & \colorbox{palered}{0.55s}\\
\textbf{$\Design$ (ours)}  & \colorbox{palered}{66.02/64.78} & \colorbox{palered}{65.58}/64.29 & \colorbox{palered}{66.01/64.78} & \colorbox{palered}{65.96}/64.62 & \colorbox{palered}{65.89/64.62} & 0.70s\\
\bottomrule
\end{tabular}
\vspace{2mm}
\caption{\textbf{BEVDepth} based UQ results on aiMotive under corner cases: motion-blur, over-exposure, under-exposure, LiDAR-noise. Metrics: all-point AP / 11-point interpolation AP (\%). The best (second-best) results are highlighted in \colorbox{palered}{pink} (\colorbox{lavender}{purple}).}
\label{table:uq1}
\vspace{-4mm}
\end{table*}

\begin{table*}[htbp]
\centering
\footnotesize
\begin{tabular}{l c c c c c c | c}
\toprule
\textbf{UQ Method} & \textbf{Motion-Blur} & \textbf{Over-Exposed} & \textbf{Under-Exposed} & \textbf{LiDAR-Jitter} & \textbf{\begin{tabular}[c]{@{}c@{}}Event \\ Low-Resolution \end{tabular}} & \textbf{Mean} & \textbf{\begin{tabular}[c]{@{}c@{}}Inference \\ Latency \end{tabular}}\\
\midrule
PostNet~\cite{charpentier2020posterior} & 42.82/53.91 & 43.77/51.74 & 29.98/33.13 & 47.95/52.44 & 48.08/56.23 & 42.41/47.86 & $-$\\
InfMCD~\cite{charpentier2020posterior}  & 45.29/56.25  & 44.53/52.85  & 30.14/\colorbox{lavender}{36.06}  & 48.76/54.06  & 49.50/57.30  & 43.64/51.30 &0.58s\\
InfNoise~\cite{mi2022training}  & \colorbox{lavender}{47.31/56.93}  & 44.66/52.71  & \colorbox{lavender}{31.73}/34.81 & 48.79/\colorbox{lavender}{54.79}  & 50.01/\colorbox{lavender}{57.97}  & \colorbox{lavender}{44.50/51.44} & \colorbox{lavender}{0.53s}\\
LDU~\cite{franchi2022latent}  & 46.01/56.35  & \colorbox{lavender}{45.00/52.96}  & 30.38/33.46  & \colorbox{lavender}{48.98}/53.19  & \colorbox{lavender}{50.47}/57.95  & 44.17/50.78 &\colorbox{palered}{0.47s}\\
\textbf{$\Design$ (ours)}  & \colorbox{palered}{49.94/58.06}  & \colorbox{palered}{46.68/55.32}  & \colorbox{palered}{33.97/37.58}  & \colorbox{palered}{50.83/56.19}  & \colorbox{palered}{52.68/59.56}  & \colorbox{palered}{46.39/53.54} &0.64s\\
\bottomrule
\vspace{1pt}
\end{tabular}
\caption{\textbf{CMNeXt} based UQ results on DeLiVER test set (out-of-distribution data)/validation set (in-distribution data) with corner cases: motion-blur, over-exposure, under-exposure, LiDAR-jitter, and event low-resolution. Metrics: mIoU (\%). The best (second-best) results are highlighted in \colorbox{palered}{pink} (\colorbox{lavender}{purple}).}
\label{table:uq2}
\vspace{-8mm}
\end{table*}

\subsection{Implementation Details}
\subsubsection{Model Specification}

We implemented $\Design$ with PyTorch on an Ubuntu system with four NVIDIA RTX 3090 GPUs. 
\textcolor{black}{
The VLM was fine-tuned with abnormal data samples (e.g., night and snow) from the training splits of aiMotive~\cite{matuszka2022aimotive}, DeLiVER~\cite{zhang2023delivering} (excluding OOD samples), and MFNet~\cite{ha2017mfnet}. 
These training splits were also used for modality-specific feature extraction and feature-pool construction, while the simulated OOD samples were used only during inference for robustness evaluation.
}
Since our workflow is 3D-based, but DeLiVER and MFNet contain 2D images, we performed a 3D-to-2D migration by extracting features from the 2D datasets (e.g., DeLiVER and MFNet) using ResNet-50~\cite{koonce2021resnet} to build a feature pool for alignment with our framework.
\textcolor{black}{
For feature-pool construction, we set $N'=0.25N$, selecting 25\% of the training feature representations for each modality.
}
We train our model on aiMotive with the AdamW optimizer~\cite{loshchilov2017decoupled} with the learning rate of $0.0000625$ and weight decay at $10^{-6}$, 
on DeLiVER using the AdamW optimizer~\cite{loshchilov2017decoupled} with the learning rate of $0.00006$ and weight decay at $0.01$, 
and on MFNet~\cite{ha2017mfnet} with the momentum SGD~\cite{gower2019sgd} with the learning rate of $0.01$ and decayed by a factor of $0.94$ per epoch.
Each baseline was trained with a batch size of 1 for 200 epochs or until convergence.
We set the parameters $\nu$ to 1 and $\gamma$ to 0.5 in the weighting function (equation (\ref{eq:weight_1})), and both parameters $\alpha$ and $\beta$ to 1 in the loss function (equation (\ref{eq:total_loss})). We set the threshold $\tau$ in the UQ loss function (equation (\ref{eq:q_loss}), equation (\ref{eq:r_loss})) to 0.5. Our method is compared against SOTA UQ techniques and multi-modal fusion works, following each method’s open-source implementation and hyperparameter settings. 
\textcolor{black}{
For the inference latency reported in Tables~\ref{table:fusion3}--\ref{table:uq2}, $\Design$ was evaluated on a single NVIDIA RTX 3090 with a batch size of 1, using FP32 for the perception model and FP16 for the VLM.
The overall inference latency ranges from $0.64$ to $0.70$ s per frame.
The perception model requires approximately $1.2$ TFLOPs per frame, while the VLM requires approximately $3.0$ TFLOPs per modality.
The comparison between the test sample's feature representation and the representative features in the feature pool requires less than $0.05$ TFLOPs.
}



\subsubsection{OOD Sample Generation}
Each model is trained on in-distribution (ID) training data and evaluated on simulated OOD samples to assess generalization and robustness under high-uncertainty conditions. 
We simulated OOD samples for aiMotive and MFNet, as DeLiVER inherently includes a separate OOD set. 
Camera OOD samples are generated via image transformations such as Gaussian blur and exposure shifts~\cite{van2014scikit, paszke2017automatic}. 
For LiDAR, we apply random point dropping, noise injection, and partial range removal to mimic incomplete sampling, sensor errors, and occlusions~\cite{bijelic2020seeing}.
\textcolor{black}{
Among these perturbations, partial range removal affects only part of the LiDAR input and is used to mimic sensor occlusions.
}
For radar in aiMotive, we convert JSON-formatted data into point clouds and apply the same perturbations. 
Examples of our simulated OOD samples are provided in Fig.~\ref{fig:simulate}. 

\subsubsection{VLM Fine-tuning}
We fine-tune the VLM~\cite{gu2023anomalyagpt} using LoRA with parameters set to
$\textit{lora}_r=32$, $\textit{lora}_\alpha=32$, and $\textit{lora}_\textit{dropout}=0.1$ for efficient parameter adaptation.
The vision stream employs a frozen ImageBind-Huge~\cite{girdhar2023imagebind} model, extracting patch-level features from layers 8, 16, 24, and 32.
Apart from the incorporation of a prompt adapter, the base AnomalyGPT model~\cite{gu2023anomalyagpt} remains largely unchanged, retaining its pretrained weights and architectural configurations. 
Fine-tuning is conducted using the AdamW optimizer~\cite{loshchilov2017decoupled} with a learning rate of $1e^{-3}$ and a one-cycle cosine learning rate schedule for stable convergence.
For the uncertainty heatmap (equation (\ref{eq:w_r^m})), we set $\lambda=1$ and $\gamma=0.5$ to balance the contributions from the visual and textual streams.

\subsection{Experimental Results}
\subsubsection{Multi-tasks Perception Comparison}


We compare $\Design$ against SOTA multi-modality fusion methods in Tables \ref{table:fusion1}, \ref{table:fusion2}, and \ref{table:fusion3}. 
As shown in Table \ref{table:fusion1}, on aiMotive, $\Design$ delivers on average $3.82\%$ better performance in all-point AP and $3.56\%$ in 11-point interpolation AP compared to its baseline (aiMotive) without UQ. 
It also consistently outperforms task-specific SOTA sensor fusion methods, e.g., BEVFusion~\cite{liu2022bevfusion} and MSMDFusion~\cite{Jiao_2023_CVPR}.
Note that our model-agnostic UQ module can be integrated into other SOTA fusion frameworks to enhance performance. 
As shown in Table \ref{table:fusion2}, $\Design$ exhibits better  performance compared to SOTA 
sensor fusion frameworks across all the scenarios on DeLiVER, achieving on average $6.71\%$ higher mIoU on the test set and $4.41\%$ higher mIoU on the validation set over its baseline (CMNeXt) without UQ.
\textcolor{black}{It also outperforms other SOTA fusion framework by on average 5.95\% higher mIoU on the test dataset and 3.48\% higher mIoU on the validation dataset.}
\textcolor{black}{On the MFNet (Table \ref{table:fusion3}), $\Design$ delivers on average $1.56\%$ higher mIoU over the second-best result.} 
These results demonstrate $\Design$'s better generalization in various OOD scenarios over SOTA approaches.

\begin{figure}[t]
\centering
\includegraphics[width=0.95\linewidth]{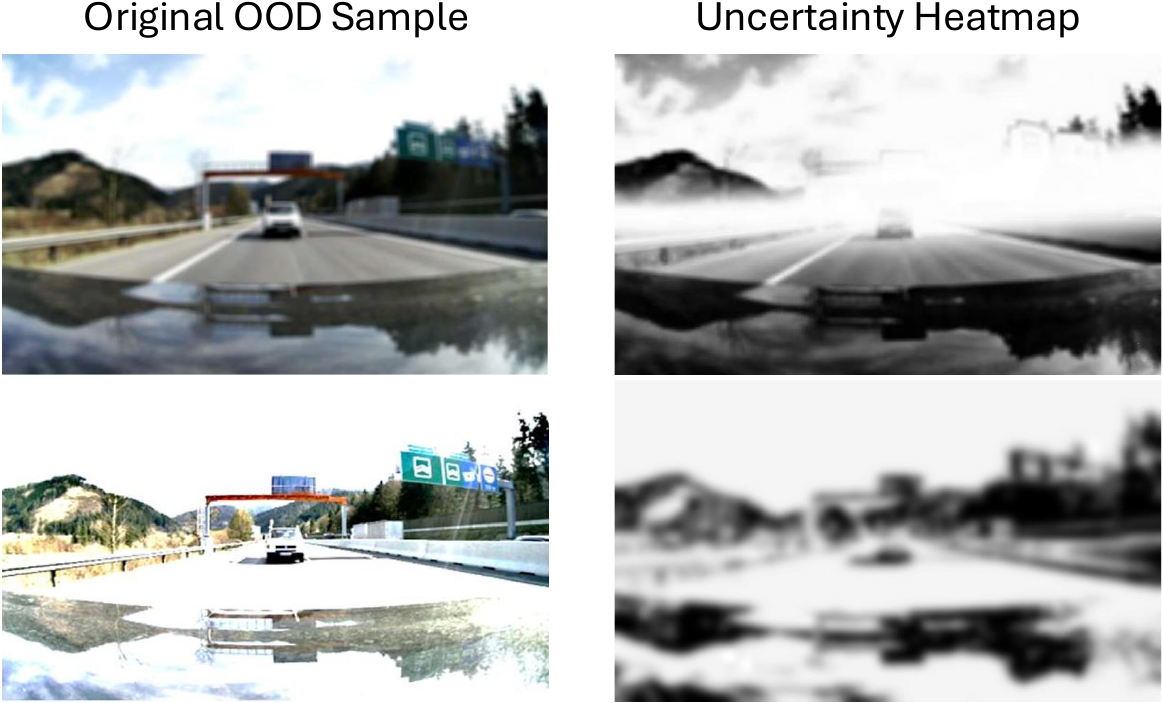}
\caption{Visualization of uncertainty heatmaps ($W^m_r$) generated by the VLM-guided UQ module for OOD samples. The figure shows uncertainty heatmaps for two camera OOD samples (motion blur and over-exposure). Brighter areas in the heatmap correspond to high-uncertainty regions, effectively localizing the visually degraded parts of the image.}
   \label{fig:heatmap}
\vspace{-5mm}
\end{figure}


\subsubsection{UQ Methods Comparison}
Tables \ref{table:fusion3}, \ref{table:uq1} and \ref{table:uq2} present $\Design$'s performance and inference latency across various OOD scenarios on MFNet, aiMotive and DeLiVER, respectively. 
$\Design$ outperforms the second-best model by on average $2.98\%$ in all-point AP and $2.20\%$ in 11-point interpolation AP on aiMotive, $1.56\%$ on MFNet, and $4.24\%$ higher mIoU on the testing set and $4.08\%$ higher mIoU on the validation set of DeLiVER. 
\textcolor{black}{
These results demonstrate that our approach provides effective UQ with limited additional computational overhead. 
The VLM and Prompt Adapter each introduce only $0.01$--$0.02$ s of additional inference latency, indicating that the semantic reasoning contributes only a small portion of the overall inference time.
}

\subsubsection{Qualitative Result}
Figure~\ref{fig:heatmap} demonstrates the uncertainty heatmaps produced by the VLM module for out-of-distribution (OOD) inputs. 
\textcolor{black}{
Although motion blur and over-exposure are applied to the whole camera image, the VLM assigns higher uncertainty to regions more affected by degradation while maintaining lower uncertainty in well-structured areas.
}
Such fine-grained uncertainty estimation enables the fusion module to down-weight unreliable signals and improve the robustness of downstream perception tasks.

\subsection{Ablation Study}
\subsubsection{Core Sampling Algorithm}
As demonstrated in Table \ref{tab:computation1}, we compare the performance and inference latency of our core sampling algorithm with using other sampling algorithms~\cite{van2017neural, kulesza2012determinantal,kokot2017kmc} for constructing the feature pool. Across all datasets, \textcolor{black}{$\Design$ outperforms determinantal point processes (DPP) by $0.28\%$ while improving inference latency by $7.6\%$.} It also outperforms vector quantised-variational autoencoder (VQ-VAE) method by $3.67\%$ with minimal increase in latency overhead. These results demonstrate that our core sampling approach effectively preserves the diversity of training samples while maintaining computational efficiency.

\begin{table}[!ht]
\footnotesize
\centering
    \addtolength{\tabcolsep}{-0.2em}{
        \begin{tabular}{l|ccc}
            \toprule
            \multirow{2}{1em}{\textbf{Method}} & {aiMotive}~\cite{matuszka2023aimotive} 
            &  
           DeLiVER~\cite{zhang2023delivering}
            & 
           MFNet~\cite{ha2017mfnet}
            \\
            &\scriptsize{(all-point AP/latency)} &  \multicolumn{2}{c}{\scriptsize{(mIoU/latency)}} \\
            \midrule
            {VQ-VAE~\cite{van2017neural}} &63.15/\colorbox{palered}{0.64s}  &44.08/\colorbox{palered}{0.61s}  & 57.96/\colorbox{palered}{0.63s}\\
            {KMC~\cite{kokot2017kmc}} &{64.76}/\colorbox{lavender}{0.65s} &\colorbox{lavender}{45.46}/\colorbox{palered}{0.61s} & 59.13/\colorbox{lavender}{0.64s}\\
           {DPP~\cite{kulesza2012determinantal}} &\colorbox{lavender}{65.73}/0.76s &46.25/0.69s & \colorbox{lavender}{59.57}/0.69s\\
            \makecell[l]{{$\Design$ (ours)}} & \colorbox{palered}{65.89}/0.70s & \colorbox{palered}{46.39}/\colorbox{lavender}{0.64s} & \colorbox{palered}{59.78}/0.66s\\
            \bottomrule
        \end{tabular}
    }
   \vspace{2mm}
    \caption{Comparison of different core sampling algorithm.}
    \label{tab:computation1}
   \vspace{-4mm}
\end{table}

\subsubsection{VLM-Guided Uncertainty Quantification}

Table \ref{tab:computation2} clearly demonstrates the substantial benefits of incorporating the VLM and the prompt adapter within the $\Design$ framework. The baseline comparison, the  ``w/o VLM'' variant, is constructed to isolate the VLM's contribution: it retains the full downstream feature enhancement and dynamic adaptation pipeline, but critically replaces the refined VLM-guided uncertainty heatmap $W^m$ with the raw, initial uncertainty estimate $Q_m$. This controlled setup ensures a fair comparison by rigorously evaluating the quality of the UQ signal alone. Across all datasets and perception tasks, integrating the VLM leads to a substantial and consistent performance improvement, with an average increase of $3.42\%$ compared to the variant without it. This result emphatically highlights the VLM's pivotal role in transforming simple distance-based estimates ($Q_m$) into  context-aware, fine-grained uncertainty maps ($W^m$), thereby significantly enhancing uncertainty estimation and subsequent feature fusion. Furthermore, the act of explicitly conditioning the VLM using the Prompt Adapter further improves performance by an additional $0.81\%$. This incremental gain validates the Prompt Adapter's effectiveness in refining the VLM's uncertainty reasoning and providing the necessary contextual and positional understanding to the textual stream. 
\textcolor{black}{
Notably, incorporating the VLM increases the overall inference latency by $0.01$--$0.02$ s, while the Prompt Adapter introduces a further $0.01$--$0.02$ s.
}

\begin{table}[!ht]
\footnotesize
\centering
    \addtolength{\tabcolsep}{-0em}{
        \begin{tabular}{l|ccc}
            \toprule
            \textbf{Methods} & \begin{tabular}[c]{@{}c@{}}{\Design} \\ {w/o VLM} \end{tabular}
            &  
           \begin{tabular}[c]{@{}c@{}}{\Design} \\ \scriptsize{w/o prompt adaptor} \end{tabular}
            & 
           \begin{tabular}[c]{@{}c@{}}{\Design} \\ \scriptsize{w prompt adaptor} \end{tabular}
            \\
            \midrule
            {aiMotive~\cite{matuszka2023aimotive}} &63.56/\colorbox{palered}{0.67s}  &\colorbox{lavender}{65.28}/\colorbox{lavender}{0.69s}  & \colorbox{palered}{65.89}/0.70s\\
            {DELIVER~\cite{zhang2023delivering}} &44.52/\colorbox{palered}{0.61s} &\colorbox{lavender}{46.08}/\colorbox{lavender}{0.63s} & \colorbox{palered}{46.39}/0.64s\\
           {MFNet~\cite{ha2017mfnet}} &58.66/\colorbox{palered}{0.63s} &\colorbox{lavender}{59.27}/\colorbox{lavender}{0.64s} & \colorbox{palered}{59.78}/0.66s\\
            \bottomrule
        \end{tabular}
    }
   \vspace{1mm}
    \caption{Comparison of the effectiveness of our VLM and prompt adapter (Metrics: mIoU/Inference Latency for DeLiVER and MFNet and all-point AP/ Inference Latency for aiMotive).}
    \label{tab:computation2}
   \vspace{-4mm}
\end{table}

\subsubsection{Impact of Fusion Strategy}
 We evaluate the effectiveness of the spatial-, channel-, and pixel-level attention along the dynamic adaptation mechanism in Table \ref{table:abla3}, using aiMotive under both in-distribution (ID) and simulated OOD scenarios. The results demonstrate a consistent performance improvement as more fusion mechanisms are incorporated. 
When all components are combined, our approach improves all-point and 11-point AP by 7.53\% and 7.07\%, respectively, under OOD conditions, and by 7.58\%  and 7.36\%, respectively, under the ID conditions over the baseline, validating its effectiveness in improving robustness and generalization. 

\begin{table}[htbp]
\footnotesize
\centering
 \addtolength{\tabcolsep}{-0.2em}{\begin{tabular}{c c | c | c c}
\toprule
\multicolumn{2}{c|}{\textbf{Feature Enhancement}} & Dynamic & \multicolumn{2}{c}{\textbf{All-point / 11-point AP (\%)}} \\
\cmidrule(lr){1-2} \cmidrule(lr){4-5}
\textbf{Spatial/Channel} & \textbf{Pixel} & Adaptation & \textbf{ID} & \textbf{OOD} \\
\midrule
\xmark & \xmark & \xmark & 65.84 / 64.47 & 61.32 / 59.85 \\
\cmark & \xmark & \xmark & 67.53 / 66.05 & 62.74 / 61.12 \\
\xmark & \cmark & \xmark & 67.82 / 66.23 & 63.08 / 61.43 \\
\cmark & \cmark & \xmark & 69.31 / 67.64 & 64.25 / 62.63 \\
\xmark & \xmark & \cmark & 71.67 / 70.03 & 66.42 / 64.74 \\
\cmark & \cmark & \cmark & \textbf{73.42} / \textbf{71.83} & \textbf{68.85} / \textbf{66.92} \\
\bottomrule
\end{tabular}}
\vspace{2mm}
\caption{Impact of the attention mechanisms and dynamic adaptation mechanism in the fusion module on aiMotive}
\label{table:abla3}
\vspace{-8mm}
\end{table}



\section{Conclusion}\label{sec:conclusion}
We present $\Design$, a novel uncertainty-aware multi-modal sensor fusion framework that significantly improves the robustness and generalization capabilities of AV perception systems. 
Our primary innovation lies in the effective combination of two key components: VLM-guided uncertainty estimation and a dynamic adaptation mechanism. 
The VLM module leverages extensive prior knowledge to transform coarse reliability scores into fine-grained, context-aware uncertainty heatmaps. 
\textcolor{black}{By combining local feature differences with global contextual reasoning, the VLM is expected to support uncertainty quantification in novel semantic scenes.}
The dynamic adaptation mechanism is specifically designed to explicitly model and capture cross-modal dependencies, \textcolor{black}{allowing the framework to adaptively balance contributions from various sensors and retain useful information from less affected modalities to support graceful degradation.}
Through these integrated innovations, $\Design$ successfully mitigates uncertainty propagation and achieves superior performance in complex, OOD scenarios. 
\textcolor{black}{Future work will investigate how to identify unreliable VLM guidance and limit its influence on fusion in extreme semantic edge cases.}

\bibliographystyle{IEEEtran}
\bibliography{reference}

@String(CVPR= {IEEE Conf. Comput. Vis. Pattern Recog.})

@String(ICCV= {Int. Conf. Comput. Vis.})

@String(ICASSP=	{ICASSP})

@String(ICLR = {Int. Conf. Learn. Represent.})

@String(AAAI = {AAAI})

@String(CVPR  = {CVPR})

@String(ICCV  = {ICCV})

@String(ICLR  = {ICLR})

@inproceedings{matuszka2023aimotive,
title={aiMotive Dataset: A Multimodal Dataset for Robust Autonomous Driving with Long-Range Perception},
author={Tamas Matuszka},
booktitle={International Conference on Learning Representations 2023 Workshop on Scene Representations for Autonomous Driving},
year={2023},
url={https://openreview.net/forum?id=LW3bRLlY-SA}
}

@article{matuszka2022aimotive,
  title={aimotive dataset: A multimodal dataset for robust autonomous driving with long-range perception},
  author={Matuszka, Tam{\'a}s and others},
  journal={arXiv preprint arXiv:2211.09445},
  year={2022}
}

@inproceedings{zhang2023delivering,
  title={Delivering arbitrary-modal semantic segmentation},
  author={Zhang, Jiaming and others},
  booktitle={Proceedings of the IEEE/CVF Conference on Computer Vision and Pattern Recognition},
  pages={1136--1147},
  year={2023}
}

@inproceedings{ha2017mfnet,
  title={MFNet: Towards real-time semantic segmentation for autonomous vehicles with multi-spectral scenes},
  author={Ha, Qishen and others},
  booktitle={International Conference on Intelligent Robots and Systems}, 
  year={2017},
  organization={IEEE}
}

@InProceedings{Jiao_2023_CVPR,
    author    = {Jiao, Yang and Jie, Zequn and Chen, Shaoxiang and Chen, Jingjing and Ma, Lin and Jiang, Yu-Gang},
    title     = {MSMDFusion: Fusing LiDAR and Camera at Multiple Scales With Multi-Depth Seeds for 3D Object Detection},
    booktitle = {Proceedings of the IEEE/CVF Conference on Computer Vision and Pattern Recognition (CVPR)}, 
    year      = {2023}}

@InProceedings{focalformer3d,
  title={FocalFormer3D: Focusing on Hard Instance for 3D Object Detection},
  author={Chen, Yilun and others},
  journal={ICCV},
  year={2023}
}

@article{yan2023cross,
  title={Cross Modal Transformer via Coordinates Encoding for 3D Object Dectection},
  author={Yan, Junjie and others},
  journal={arXiv preprint arXiv:2301.01283},
  year={2023}
}

@inproceedings{liu2022bevfusion,
  title={BEVFusion: Multi-Task Multi-Sensor Fusion with Unified Bird's-Eye View Representation},
  author={Liu, Zhijian and others},
  booktitle={ICRA},
  year={2023}
}

@inproceedings{mi2022training,
  title={Training-free uncertainty estimation for dense regression: Sensitivity as a surrogate},
  author={Mi, Lu and Wang, Hao and Tian, Yonglong and He, Hao and Shavit, Nir N},
  booktitle={Proceedings of the AAAI Conference on Artificial Intelligence}, 
  year={2022}
}

@inproceedings{franchi2022latent,
  title={Latent discriminant deterministic uncertainty},
  author={Franchi, Gianni and Yu, Xuanlong and Bursuc, Andrei and Aldea, Emanuel and Dubuisson, Severine and Filliat, David},
  booktitle={European Conference on Computer Vision},
  pages={243--260},
  year={2022},
  organization={Springer}
}

@article{li2024stitchfusion,
  title={Stitchfusion: Weaving any visual modalities to enhance multimodal semantic segmentation},
  author={Li, Bingyu and Zhang, Da and Zhao, Zhiyuan and Gao, Junyu and Li, Xuelong},
  journal={arXiv preprint arXiv:2408.01343},
  year={2024}
}

@inproceedings{wang2022multimodal,
  title={Multimodal token fusion for vision transformers},
  author={Wang, Yikai and Chen, Xinghao and Cao, Lele and Huang, Wenbing and Sun, Fuchun and Wang, Yunhe},
  booktitle={Proceedings of the IEEE/CVF conference on computer vision and pattern recognition},
  year={2022}
}

@inproceedings{broedermann2023hrfuser,
  title={HRFuser: A multi-resolution sensor fusion architecture for 2D object detection},
  author={Broedermann, Tim and others},
  booktitle={International Conference on Intelligent Transportation Systems}, 
  year={2023},
  organization={IEEE}
}

@article{charpentier2020posterior,
  title={Posterior network: Uncertainty estimation without ood samples via density-based pseudo-counts},
  author={Charpentier, Bertrand and Z{\"u}gner, Daniel and G{\"u}nnemann, Stephan},
  journal={Advances in neural information processing systems}, 
  year={2020}
}

@inproceedings{liu2021swin,
  title={Swin transformer: Hierarchical vision transformer using shifted windows},
  author={Liu, Ze and Lin, Yutong and Cao, Yue and Hu, Han and Wei, Yixuan and Zhang, Zheng and Lin, Stephen and Guo, Baining},
  booktitle={Proceedings of the IEEE/CVF international conference on computer vision},
  pages={10012--10022},
  year={2021}
}

@article{xie2021segformer,
  title={SegFormer: Simple and efficient design for semantic segmentation with transformers},
  author={Xie, Enze and Wang, Wenhai and Yu, Zhiding and Anandkumar, Anima and Alvarez, Jose M and Luo, Ping},
  journal={Advances in neural information processing systems},
  volume={34},
  pages={12077--12090},
  year={2021}
}

@inproceedings{hu2019acnet,
  title={Acnet: Attention based network to exploit complementary features for rgbd semantic segmentation},
  author={Hu, Xinxin and Yang, Kailun and Fei, Lei and Wang, Kaiwei},
  booktitle={International conference on image processing}, 
  year={2019},
  organization={IEEE}
}

@inproceedings{deng2021feanet,
  title={FEANet: Feature-enhanced attention network for RGB-thermal real-time semantic segmentation},
  author={Deng, Fuqin and Feng, Hua and Liang, Mingjian and Wang, Hongmin and Yang, Yong and Gao, Yuan and Chen, Junfeng and Hu, Junjie and Guo, Xiyue and Lam, Tin Lun},
  booktitle={International conference on intelligent robots and systems (IROS)},
  year={2021},
  organization={IEEE}
}

@article{sun2020fuseseg,
  title={FuseSeg: Semantic segmentation of urban scenes based on RGB and thermal data fusion},
  author={Sun, Yuxiang and Zuo, Weixun and Yun, Peng and Wang, Hengli and Liu, Ming},
  journal={Transactions on Automation Science and Engineering},
  year={2020},
  publisher={IEEE}
}

@article{zhang2023cmx,
  title={CMX: Cross-modal fusion for RGB-X semantic segmentation with transformers},
  author={Zhang, Jiaming and Liu, Huayao and Yang, Kailun and Hu, Xinxin and Liu, Ruiping and Stiefelhagen, Rainer},
  journal={IEEE Transactions on Intelligent Transportation Systems},
  year={2023}
}

@inproceedings{liu2023segmif,
  title={Multi-interactive Feature Learning and a Full-time Multi-modality Benchmark for Image Fusion and Segmentation},
  author={Liu, Jinyuan and Liu, Zhu and Wu, Guanyao and Ma, Long and Liu, Risheng and Zhong, Wei and Luo, Zhongxuan and Fan, Xin},
  booktitle={International Conference on Computer Vision},
  year={2023}
}

@article{gu2023anomalyagpt,
  title={AnomalyGPT: Detecting Industrial Anomalies using Large Vision-Language Models},
  author={Gu, Zhaopeng and Zhu, Bingke and Zhu, Guibo and Chen, Yingying and Tang, Ming and Wang, Jinqiao},
  journal={arXiv preprint arXiv:2308.15366},
  year={2023}
}

@article{yin2021multimodal,
  title={Multimodal virtual point 3d detection},
  author={Yin, Tianwei and Zhou, Xingyi and Kr{\"a}henb{\"u}hl, Philipp},
  journal={Advances in Neural Information Processing Systems}, 
  year={2021}
}

@inproceedings{roth2022towards,
  title={Towards total recall in industrial anomaly detection},
  author={Roth, Karsten and others},
  booktitle={CVPR},
  year={2022}
}

@inproceedings{bijelic2020seeing,
  title={Seeing through fog without seeing fog: Deep multimodal sensor fusion in unseen adverse weather},
  author={Bijelic, Mario and others},
  booktitle={IEEE/CVF Conference on Computer Vision and Pattern Recognition},
  year={2020}
}

@article{feng2020deep,
  title={Deep multi-modal object detection and semantic segmentation for autonomous driving: Datasets, methods, and challenges},
  author={Feng, Di and Haase-Sch{\"u}tz, Christian and Rosenbaum, Lars and Hertlein, Heinz and Glaeser, Claudius and Timm, Fabian and Wiesbeck, Werner and Dietmayer, Klaus},
  journal={IEEE Transactions on Intelligent Transportation Systems},
  volume={22},
  number={3},
  pages={1341--1360},
  year={2020},
  publisher={IEEE}
}

@article{lakshminarayanan2017simple,
  title={Simple and scalable predictive uncertainty estimation using deep ensembles},
  author={Lakshminarayanan, Balaji and Pritzel, Alexander and Blundell, Charles},
  journal={Advances in neural information processing systems},
  volume={30},
  year={2017}
}

@article{kwon2020uncertainty,
  title={Uncertainty quantification using Bayesian neural networks in classification: Application to biomedical image segmentation},
  author={Kwon, Yongchan and Won, Joong-Ho and Kim, Beom Joon and Paik, Myunghee Cho},
  journal={Computational Statistics \& Data Analysis},
  volume={142},
  pages={106816},
  year={2020},
  publisher={Elsevier}
}

@inproceedings{gal2016dropout,
  title={Dropout as a bayesian approximation: Representing model uncertainty in deep learning},
  author={Gal, Yarin and Ghahramani, Zoubin},
  booktitle={international conference on machine learning},
  pages={1050--1059},
  year={2016},
  organization={PMLR}
}

@inproceedings{bai2022transfusion,
  title={Transfusion: Robust lidar-camera fusion for 3d object detection with transformers},
  author={Bai, Xuyang and Hu, Zeyu and Zhu, Xinge and Huang, Qingqiu and Chen, Yilun and Fu, Hongbo and Tai, Chiew-Lan},
  booktitle={Proceedings of the IEEE/CVF conference on computer vision and pattern recognition},
  pages={1090--1099},
  year={2022}
}

@article{lu2023multi,
  title={The multi-modal fusion in visual question answering: a review of attention mechanisms},
  author={Lu, Siyu and Liu, Mingzhe and Yin, Lirong and Yin, Zhengtong and Liu, Xuan and Zheng, Wenfeng},
  journal={PeerJ Computer Science},
  volume={9},
  pages={e1400},
  year={2023},
  publisher={PeerJ Inc.}
}

@article{fayyad2020deep,
  title={Deep learning sensor fusion for autonomous vehicle perception and localization: A review},
  author={Fayyad, Jamil and Jaradat, Mohammad A and Gruyer, Dominique and Najjaran, Homayoun},
  journal={Sensors},
  volume={20},
  number={15},
  pages={4220},
  year={2020},
  publisher={MDPI}
}

@inproceedings{bhupathiraju2023emi,
  title={EMI-LiDAR: Uncovering Vulnerabilities of LiDAR Sensors in Autonomous Driving Setting Using Electromagnetic Interference},
  author={Bhupathiraju, Sri Hrushikesh Varma and Sheldon, Jennifer and Bauer, Luke A and Bindschaedler, Vincent and Sugawara, Takeshi and Rampazzi, Sara},
  booktitle={Proceedings of the 16th ACM Conference on Security and Privacy in Wireless and Mobile Networks},
  pages={329--340},
  year={2023}
}

@article{li2020lidar,
  title={Lidar for autonomous driving: The principles, challenges, and trends for automotive lidar and perception systems},
  author={Li, You and Ibanez-Guzman, Javier},
  journal={IEEE Signal Processing Magazine},
  year={2020},
  publisher={IEEE}
}

@inproceedings{chen2023futr3d,
  title={Futr3d: A unified sensor fusion framework for 3d detection},
  author={Chen, Xuanyao and Zhang, Tianyuan and Wang, Yue and Wang, Yilun and Zhao, Hang},
  booktitle={IEEE/CVF conference on computer vision and pattern recognition},
  year={2023}
}

@article{rahaman2021uncertainty,
  title={Uncertainty quantification and deep ensembles},
  author={Rahaman, Rahul and others},
  journal={Advances in neural information processing systems}, 
  year={2021}
}

@inproceedings{xu2020squeezesegv3,
  title={Squeezesegv3: Spatially-adaptive convolution for efficient point-cloud segmentation},
  author={Xu, Chenfeng and Wu, Bichen and Wang, Zining and Zhan, Wei and Vajda, Peter and Keutzer, Kurt and Tomizuka, Masayoshi},
  booktitle={Computer Vision--ECCV 2020: 16th European Conference, Glasgow, UK, August 23--28, 2020, Proceedings, Part XXVIII 16},
  pages={1--19},
  year={2020},
  organization={Springer}
}

@incollection{lou2023uncertainty,
  title={Uncertainty-Encoded Multi-Modal Fusion for Robust Object Detection in Autonomous Driving},
  author={Lou, Yang and Song, Qun and Xu, Qian and Tan, Rui and Wang, Jianping},
  booktitle={ECAI 2023},
  pages={1593--1600},
  year={2023},
  publisher={IOS Press}
}

@inproceedings{sander2013bayesian,
  title={Bayesian fusion: Modeling and application},
  author={Sander, Jennifer and Beyerer, Jurgen},
  booktitle={2013 Workshop on Sensor Data Fusion: Trends, Solutions, Applications (SDF)},
  pages={1--6},
  year={2013},
  organization={IEEE}
}

@article{brena2020choosing,
  title={Choosing the best sensor fusion method: A machine-learning approach},
  author={Brena, Ramon F and Aguileta, Antonio A and Trejo, Luis A and Molino-Minero-Re, Erik and Mayora, Oscar},
  journal={Sensors},
  volume={20},
  number={8},
  pages={2350},
  year={2020},
  publisher={MDPI}
}

@article{guan2023trustworthy,
  title={Trustworthy Sensor Fusion against Inaudible Command Attacks in Advanced Driver-Assistance Systems},
  author={Guan, Jiwei and Pan, Lei and Wang, Chen and Yu, Shui and Gao, Longxiang and Zheng, Xi},
  journal={IEEE Internet of Things Journal}, 
  year={2023},
  publisher={IEEE}
}

@inproceedings{malawade2022hydrafusion,
  title={HydraFusion: Context-aware selective sensor fusion for robust and efficient autonomous vehicle perception},
  author={Malawade, Arnav Vaibhav and Mortlock, Trier and Al Faruque, Mohammad Abdullah},
  booktitle={2022 ACM/IEEE 13th International Conference on Cyber-Physical Systems},
  pages={68--79},
  year={2022},
  organization={IEEE}
}

@inproceedings{shekhar1986sensor,
  title={Sensor fusion and object localization},
  author={Shekhar, Shashank and Khatib, Oussama and Shimojo, Makoto},
  booktitle={Proceedings. 1986 IEEE International Conference on Robotics and Automation},
  volume={3},
  pages={1623--1628},
  year={1986},
  organization={IEEE}
}

@inproceedings{xu2018multi,
  title={Multi-level fusion based 3d object detection from monocular images},
  author={Xu, Bin and Chen, Zhenzhong},
  booktitle={Proceedings of the IEEE conference on computer vision and pattern recognition},
  pages={2345--2353},
  year={2018}
}

@article{huang2020multi,
  title={Multi-modal sensor fusion-based deep neural network for end-to-end autonomous driving with scene understanding},
  author={Huang, Zhiyu and Lv, Chen and Xing, Yang and Wu, Jingda},
  journal={IEEE Sensors Journal},
  year={2020},
  publisher={IEEE}
}

@inproceedings{malawade2022ecofusion,
  title={EcoFusion: Energy-aware adaptive sensor fusion for efficient autonomous vehicle perception},
  author={Malawade, Arnav Vaibhav and Mortlock, Trier and Faruque, Mohammad Abdullah Al},
  booktitle={ACM/IEEE Design Automation Conference}, 
  year={2022}
}

@article{loshchilov2017decoupled,
  title={Decoupled weight decay regularization},
  author={Loshchilov, I},
  journal={arXiv preprint arXiv:1711.05101},
  year={2017}
}

@inproceedings{gower2019sgd,
  title={SGD: General analysis and improved rates},
  author={Gower, Robert Mansel and others},
  booktitle={International conference on machine learning}, 
  year={2019},
  organization={PMLR}
}

@inproceedings{paszke2017automatic,
  title={Automatic differentiation in PyTorch},
  author={Paszke, Adam and others},
  booktitle={NIPS-W},
  year={2017}
}

@article{van2014scikit,
  title={scikit-image: image processing in Python},
  author={Van der Walt, Stefan and Sch{\"o}nberger, Johannes L and Nunez-Iglesias, Juan and Boulogne, Fran{\c{c}}ois and Warner, Joshua D and Yager, Neil and Gouillart, Emmanuelle and Yu, Tony},
  journal={PeerJ},
  volume={2},
  pages={e453},
  year={2014},
  publisher={PeerJ Inc.}
}

@article{jung2023beyond,
  title={Beyond unimodal: Generalising neural processes for multimodal uncertainty estimation},
  author={Jung, Myong Chol and Zhao, He and Dipnall, Joanna and Du, Lan},
  journal={Advances in Neural Information Processing Systems}, 
  year={2023}
}

@article{liang2023quantifying,
  title={Quantifying \& modeling multimodal interactions: An information decomposition framework},
  author={Liang, Paul Pu and others},
  journal={Advances in Neural Information Processing Systems}, 
  year={2023}
}

@inproceedings{zhuang2021perception,
  title={Perception-aware multi-sensor fusion for 3d lidar semantic segmentation},
  author={Zhuang, Zhuangwei and Li, Rong and Jia, Kui and Wang, Qicheng and Li, Yuanqing and Tan, Mingkui},
  booktitle={IEEE/CVF international conference on computer vision},
  year={2021}
}

@inproceedings{meyer2019sensor,
  title={Sensor fusion for joint 3d object detection and semantic segmentation},
  author={Meyer, Gregory P and Charland, Jake and Hegde, Darshan and Laddha, Ankit and Vallespi-Gonzalez, Carlos},
  booktitle={Proceedings of the IEEE/CVF conference on computer vision and pattern recognition workshops}, 
  year={2019}
}

@article{mukherjee2021decentralized,
  title={A decentralized sensor fusion scheme for multi sensorial fault resilient pose estimation},
  author={Mukherjee, Moumita and others},
  journal={Sensors},
  year={2021},
  publisher={MDPI}
}

@article{zhu2013variational,
  title={A variational Bayesian approach to robust sensor fusion based on Student-t distribution},
  author={Zhu, Hao and Leung, Henry and He, Zhongshi},
  journal={Information Sciences},
  volume={221},
  pages={201--214},
  year={2013},
  publisher={Elsevier}
}

@article{koonce2021resnet,
  title={ResNet 50},
  author={Koonce, Brett and others},
  journal={Convolutional neural networks with swift for tensorflow: image recognition and dataset categorization},
  year={2021},
  publisher={Springer}
}

@inproceedings{zhou2018voxelnet,
  title={Voxelnet: End-to-end learning for point cloud based 3d object detection},
  author={Zhou, Yin and Tuzel, Oncel},
  booktitle={Proceedings of the IEEE conference on computer vision and pattern recognition},
  pages={4490--4499},
  year={2018}
}

@article{wang2023large,
  title={Large kernel sparse ConvNet weighted by multi-frequency attention for remote sensing scene understanding},
  author={Wang, Junjie and others},
  journal={IEEE Transactions on Geoscience and Remote Sensing},
  year={2023},
  publisher={IEEE}
}

@inproceedings{radford2021learning,
  title={Learning transferable visual models from natural language supervision},
  author={Radford, Alec and Kim, Jong Wook and Hallacy, Chris and Ramesh, Aditya and Goh, Gabriel and Agarwal, Sandhini and Sastry, Girish and Askell, Amanda and Mishkin, Pamela and Clark, Jack and others},
  booktitle={International conference on machine learning},
  pages={8748--8763},
  year={2021},
  organization={PmLR}
}

@inproceedings{li2023blip,
  title={Blip-2: Bootstrapping language-image pre-training with frozen image encoders and large language models},
  author={Li, Junnan and others},
  booktitle={International conference on machine learning},
  pages={19730--19742},
  year={2023},
  organization={PMLR}
}

@article{zang2024overcoming,
  title={Overcoming the pitfalls of vision-language model finetuning for OOD generalization},
  author={Zang, Yuhang and Goh, Hanlin and Susskind, Josh and Huang, Chen},
  journal={arXiv preprint arXiv:2401.15914},
  year={2024}
}

@inproceedings{addepalli2024leveraging,
  title={Leveraging vision-language models for improving domain generalization in image classification},
  author={Addepalli, Sravanti and Asokan, Ashish Ramayee and Sharma, Lakshay and Babu, R Venkatesh},
  booktitle={Proceedings of the IEEE/CVF Conference on Computer Vision and Pattern Recognition},
  pages={23922--23932},
  year={2024}
}

@inproceedings{chen2024practicaldg,
  title={Practicaldg: Perturbation distillation on vision-language models for hybrid domain generalization},
  author={Chen, Zining and Wang, Weiqiu and Zhao, Zhicheng and Su, Fei and Men, Aidong and Meng, Hongying},
  booktitle={Proceedings of the IEEE/CVF Conference on Computer Vision and Pattern Recognition},
  pages={23501--23511},
  year={2024}
}

@article{gawlikowski2023survey,
  title={A survey of uncertainty in deep neural networks},
  author={Gawlikowski, Jakob and Tassi, Cedrique Rovile Njieutcheu and Ali, Mohsin and Lee, Jongseok and Humt, Matthias and Feng, Jianxiang and Kruspe, Anna and Triebel, Rudolph and Jung, Peter and Roscher, Ribana and others},
  journal={Artificial Intelligence Review}, 
  year={2023},
  publisher={Springer}
}

@article{van2017neural,
  title={Neural discrete representation learning},
  author={Van Den Oord, Aaron and Vinyals, Oriol and others},
  journal={Advances in neural information processing systems}, 
  year={2017}
}

@article{kulesza2012determinantal,
  title={Determinantal point processes for machine learning},
  author={Kulesza, Alex and Taskar, Ben and others},
  journal={Foundations and Trends in Machine Learning}, 
  year={2012},
  publisher={Now Publishers, Inc.}
}

@article{kokot2017kmc,
  title={KMC 3: counting and manipulating k-mer statistics},
  author={Kokot, Marek and D{\l}ugosz, Maciej and Deorowicz, Sebastian},
  journal={Bioinformatics}, 
  year={2017},
  publisher={Oxford University Press}
}

@inproceedings{girdhar2023imagebind,
  title={Imagebind: One embedding space to bind them all},
  author={Girdhar, Rohit and others},
  booktitle={CVPR},
  year={2023}
}

@article{chiang2023vicuna,
  title={Vicuna: An open-source chatbot impressing gpt-4 with 90\%* chatgpt quality, March 2023},
  author={Chiang, Wei-Lin and Li, Zhuohan and Lin, Zi and Sheng, Ying and Wu, Zhanghao and Zhang, Hao and Zheng, Lianmin and Zhuang, Siyuan and Zhuang, Yonghao and Gonzalez, Joseph E and others},
  journal={URL https://lmsys. org/blog/2023-03-30-vicuna},
  volume={3},
  number={5},
  year={2023}
}

@article{lecun1998gradient,
  title={Gradient-based learning applied to document recognition},
  author={LeCun, Yann and others},
  journal={Proceedings of the IEEE},
  publisher={Ieee}
}

@inproceedings{wang2024rs2g,
  title={Rs2g: Data-driven scene-graph extraction and embedding for robust autonomous perception and scenario understanding},
  author={Wang, Junyao and others},
  booktitle={IEEE/CVF Winter Conference on Applications of Computer Vision},
  pages={7493--7502},
  year={2024}
}

@inproceedings{lang2019pointpillars,
  title={Pointpillars: Fast encoders for object detection from point clouds},
  author={Lang, Alex H and others},
  booktitle={CVPR},
  year={2019}
}

@article{graham2017submanifold,
  title={Submanifold sparse convolutional networks},
  author={Graham, Benjamin and Van der Maaten, Laurens},
  journal={arXiv preprint arXiv:1706.01307},
  year={2017}
}

@article{hu2022lora,
  title={Lora: Low-rank adaptation of large language models.},
  author={Hu, Edward J and Shen, Yelong and Wallis, Phillip and Allen-Zhu, Zeyuan and Li, Yuanzhi and Wang, Shean and Wang, Lu and Chen, Weizhu and others},
  journal={ICLR},
  volume={1},
  number={2},
  pages={3},
  year={2022}
}

@inproceedings{chococoon,
  title={Cocoon: Robust Multi-Modal Perception with Uncertainty-Aware Sensor Fusion},
  author={Cho, Minkyoung and others},
  booktitle={ICLR},
  year={2024}
}

@inproceedings{palladin2024samfusion,
  title={Samfusion: Sensor-adaptive multimodal fusion for 3d object detection in adverse weather},
  author={Palladin, Edoardo and others},
  booktitle={European Conference on Computer Vision},
  year={2024},
  organization={Springer}
}

@article{brodermann2025cafuser,
  title={Cafuser: Condition-aware multimodal fusion for robust semantic perception of driving scenes},
  author={Br{\"o}dermann, Tim and Sakaridis, Christos and Fu, Yuqian and Van Gool, Luc},
  journal={IEEE Robotics and Automation Letters},
  year={2025},
  publisher={IEEE}
}

@inproceedings{sural2024contextualfusion,
  title={ContextualFusion: Context-based multi-sensor fusion for 3D object detection in adverse operating conditions},
  author={Sural, Shounak and Sahu, Nishad and Rajkumar, Ragunathan Raj},
  booktitle={Intelligent vehicles symposium}, 
  year={2024},
  organization={IEEE}
}

@inproceedings{wang2023disthd,
  title={Disthd: A learner-aware dynamic encoding method for hyperdimensional classification},
  author={Wang, Junyao and Huang, Sitao and Imani, Mohsen},
  booktitle={2023 60th ACM/IEEE Design Automation Conference (DAC)},
  year={2023},
  organization={IEEE}
}

@inproceedings{chen2025hyperdimensional,
  title={Hyperdimensional uncertainty quantification for multimodal uncertainty fusion in autonomous vehicles perception},
  author={Chen, Luke and Wang, Junyao and Mortlock, Trier and Khargonekar, Pramod and Al Faruque, Mohammad Abdullah},
  booktitle={2025 IEEE/CVF Conference on Computer Vision and Pattern Recognition (CVPR)},
  pages={22306--22316},
  year={2025},
  organization={IEEE}
}

@article{wang2023hyperdetect,
  title={Hyperdetect: A real-time hyperdimensional solution for intrusion detection in iot networks},
  author={Wang, Junyao and Xu, Haocheng and Achamyeleh, Yonatan Gizachew and Huang, Sitao and Al Faruque, Mohammad Abdullah},
  journal={IEEE Internet of Things Journal},
  volume={11},
  number={8},
  pages={14844--14856},
  year={2023},
  publisher={IEEE}
}

@inproceedings{wang2023domino,
  title={Domino: Domain-invariant hyperdimensional classification for multi-sensor time series data},
  author={Wang, Junyao and Chen, Luke and Al Faruque, Mohammad Abdullah},
  booktitle={2023 IEEE/ACM International Conference on Computer Aided Design (ICCAD)},
  pages={1--9},
  year={2023},
  organization={IEEE}
}

@inproceedings{wang2023late,
  title={Late breaking results: Scalable and efficient hyperdimensional computing for network intrusion detection},
  author={Wang, Junyao and Chen, Hanning and Issa, Mariam and Huang, Sitao and Imani, Mohsen},
  booktitle={2023 60th ACM/IEEE Design Automation Conference (DAC)},
  pages={1--2},
  year={2023},
  organization={IEEE}
}

@inproceedings{wang2025transformer,
  title={Transformer-based contrastive meta-learning for low-resource generalizable activity recognition},
  author={Wang, Junyao and Al Faruque, Mohammad Abdullah},
  booktitle={ICASSP 2025-2025 IEEE International Conference on Acoustics, Speech and Signal Processing (ICASSP)},
  pages={1--5},
  year={2025},
  organization={IEEE}
}

@inproceedings{wang2024smore,
  title={Smore: Similarity-based hyperdimensional domain adaptation for multi-sensor time series classification},
  author={Wang, Junyao and Al Faruque, Mohammad},
  booktitle={Proceedings of the 61st ACM/IEEE Design Automation Conference},
  pages={1--6},
  year={2024}
}

@inproceedings{chen2025unveiling,
  author    = {Chen, Zijun and Hu, Wenbo and He, Guande and
               Deng, Zhijie and Zhang, Zheng and Hong, Richang},
  title     = {Unveiling Uncertainty: A Deep Dive into Calibration
               and Performance of Multimodal Large Language Models},
  booktitle = {Proceedings of the 31st International Conference
               on Computational Linguistics},
  year      = {2025},
  month     = jan,
  pages     = {3095--3109},
  publisher = {Association for Computational Linguistics}
}

@inproceedings{guan2024hallusionbench,
  author    = {Guan, Tianrui and Liu, Fuxiao and Wu, Xiyang and
               Xian, Ruiqi and Li, Zongxia and Liu, Xiaoyu and
               Wang, Xijun and Chen, Lichang and Huang, Furong and
               Yacoob, Yaser and Manocha, Dinesh and Zhou, Tianyi},
  title     = {{HallusionBench}: An Advanced Diagnostic Suite
               for Entangled Language Hallucination and Visual
               Illusion in Large Vision-Language Models},
  booktitle = {Proceedings of the IEEE/CVF Conference on
               Computer Vision and Pattern Recognition (CVPR)},
  year      = {2024},
  month     = jun,
  pages     = {14375--14385}
}

@inproceedings{xie2025drivebench,
  author    = {Xie, Shaoyuan and Kong, Lingdong and Dong, Yuhao and
               Sima, Chonghao and Zhang, Wenwei and Chen, Qi Alfred
               and Liu, Ziwei and Pan, Liang},
  title     = {Are {VLMs} Ready for Autonomous Driving?
               An Empirical Study from the Reliability,
               Data and Metric Perspectives},
  booktitle = {Proceedings of the IEEE/CVF International
               Conference on Computer Vision (ICCV)},
  year      = {2025},
  month     = oct,
  pages     = {6585--6597}
}

@inproceedings{tian2020uno,
  title={UNO: Uncertainty-Aware Noisy-Or Multimodal Fusion for Unanticipated Input Degradation},
  author={Tian, Junjiao and Cheung, Wesley and Glaser, Nathaniel and Liu, Yen-Cheng and Kira, Zsolt},
  booktitle={2020 IEEE International Conference on Robotics and Automation (ICRA)},
  pages={5716--5723},
  year={2020},
  organization={IEEE},
  doi={10.1109/ICRA40945.2020.9197266}
}

@inproceedings{feng2020leveraging,
  title={Leveraging Uncertainties for Deep Multi-Modal Object Detection in Autonomous Driving},
  author={Feng, Di and Cao, Yifan and Rosenbaum, Lars and Timm, Fabian and Dietmayer, Klaus},
  booktitle={2020 IEEE Intelligent Vehicles Symposium (IV)},
  pages={877--884},
  year={2020},
  organization={IEEE},
  doi={10.1109/IV47402.2020.9304551}
}

@article{taparia2025vlc,
  title={VLC Fusion: Vision-Language Conditioned Sensor Fusion for Robust Object Detection},
  author={Taparia, Aditya and Ngu, Noel and Leiva, Mario and Kricheli, Joshua Shay and Corcoran, John and Bastian, Nathaniel D. and Simari, Gerardo I. and Shakarian, Paulo and Senanayake, Ransalu},
  journal={arXiv preprint arXiv:2505.12715},
  year={2025},
  doi={10.48550/arXiv.2505.12715}
}

\end{document}